%% file: main.tex
\documentclass[sigconf]{acmart}
\DeclareUnicodeCharacter{FF0C}{,}
\usepackage{subfigure}
\usepackage{balance}
\def\model{CAPS}
\AtBeginDocument{%
  \providecommand\BibTeX{{%
    \normalfont B\kern-0.5em{\scshape i\kern-0.25em b}\kern-0.8em\TeX}}}

\copyrightyear{2025}
\acmYear{2025}
\setcopyright{cc}
\setcctype{by}
\acmConference[KDD '25]{Proceedings of the 31st ACM SIGKDD Conference on Knowledge Discovery and Data Mining V.2}{August 3--7, 2025}{Toronto, ON, Canada}
\acmBooktitle{Proceedings of the 31st ACM SIGKDD Conference on Knowledge Discovery and Data Mining V.2 (KDD '25), August 3--7, 2025, Toronto, ON, Canada}
\acmDOI{10.1145/3711896.3736891}
\acmISBN{979-8-4007-1454-2/2025/08}

\begin{document}


\title{Continuous Optimization for Feature Selection with Permutation-Invariant Embedding and Policy-Guided Search}

\author{Rui Liu}
\affiliation{
  \institution{University of Kansas}
  \city{Lawrence}
  \state{KS}
  \country{USA}}
\email{rayliu@ku.edu}

\author{Rui Xie}
\affiliation{
  \institution{University of Central Florida}
  \city{Orlando}
  \state{FL}
  \country{USA}}
\email{rui.xie@ucf.edu}

\author{Zijun Yao}
\affiliation{
  \institution{University of Kansas}
  \city{Lawrence}
  \state{KS}
  \country{USA}}
\email{zyao@ku.edu}

\author{Yanjie Fu}
\affiliation{
  \institution{Arizona State University}
  \city{Tempe}
  \state{AZ}
  \country{USA}}
\email{yanjie.fu@asu.edu}

\author{Dongjie Wang$^{\dag}$}
\authornote{Corresponding author}

\affiliation{
  \institution{University of Kansas}
  \city{Lawrence}
  \state{KS}
  \country{USA}}
\email{wangdongjie@ku.edu}

\renewcommand{\shortauthors}{Rui Liu, Rui Xie, Zijun Yao, Yanjie Fu, \& Dongjie Wang.}





\input{abstract}

\maketitle

\input{introduction}

\input{problem}

\input{method}

\input{experiment}

\input{related}

\input{conclusion}

\newpage

\bibliographystyle{ACM-Reference-Format}
\balance
\bibliography{sample-base}

\end{document}

%% file: abstract.tex
\begin{abstract}
Feature selection removes redundant features to enhance both performance and computational efficiency in downstream tasks.
Existing methods often struggle to capture complex feature interactions and adapt to diverse scenarios.
Recent advances in this domain have incorporated generative intelligence to address these drawbacks by uncovering intricate relationships between features.
However, two key limitations remain: 1) embedding feature subsets in a continuous space is challenging due to permutation sensitivity, as changes in feature order can introduce biases and weaken the embedding learning process;
2) gradient-based search in the embedding space assumes convexity, which is rarely guaranteed, leading to reduced search effectiveness and suboptimal subsets.
To address these limitations, we propose a new framework that can: 1)  preserve feature subset knowledge in a continuous embedding space while ensuring permutation invariance; 2) effectively explore the embedding space without relying on strong convex assumptions.
For the first objective,  we develop an encoder-decoder paradigm to preserve feature selection knowledge into a continuous embedding space.
This paradigm captures feature interactions through pairwise relationships within the subset, removing the influence of feature order on the embedding. 
Moreover, an inducing point mechanism is introduced to accelerate pairwise relationship computations.
For the second objective, we employ a policy-based reinforcement learning  (RL) approach to guide the exploration of the embedding space.
The RL agent effectively navigates the space by balancing multiple objectives.
By prioritizing high-potential regions adaptively and eliminating the reliance on convexity assumptions, this search strategy effectively reduces the risk of converging to local optima.
Finally, we conduct extensive experiments to demonstrate the effectiveness, efficiency, robustness and explicitness of our model.
Our code and dataset are publicly accessible on GitHub\footnote{ \url{https://github.com/RayLiu1103/CAPS}}.

\end{abstract}

%% file: introduction.tex
\section{Introduction}
Feature selection eliminates redundant and irrelevant features to improve both predictive performance and computational efficiency in downstream tasks.
Despite the growing dominance of deep learning, feature selection remains indispensable in scenarios characterized by high-dimensional data, the need for interpretability, and limited resource constraints.
For instance, in healthcare, identifying key biomarkers from extensive genetic datasets enhances model transparency and supports informed decision-making.
In finance, extracting relevant features from transactional data facilitates the development of efficient and interpretable risk models.
These examples demonstrate the importance of feature selection in addressing real-world challenges, even in the era of deep learning.

Existing works can be divided into three categories:
1) filter methods~\cite{kbest,forman2003extensive,hall1999feature,yu2003feature} rank features based on specific scoring criteria.
They often select the top-K ranking features as the optimal subset;
2) wrapper methods~\cite{yang1998feature,kim2000feature,narendra1977branch,kohavi1997wrappers} iteratively generate and evaluate feature subsets.
They select the feature subset that performs best on a downstream task as the optimal one;
3) embedded methods~\cite{lasso,sugumaran2007feature} integrate a feature selection regularization term (e.g., $L_1$ regularization) into model training. 
As the model converges, irrelevant features are penalized, and the non-zero coefficients indicate the optimal subset.
While prior work has achieved notable success, it falls short in capturing complex feature interactions and adapting feature selection knowledge to diverse and dynamic scenarios.

Recently, the success of generative AI has inspired researchers to revisit the feature selection problem through a generative lens. By embedding discrete feature selection knowledge into a continuous embedding space, they not only capture complex feature interactions for generating enhanced feature subsets but also make the knowledge more adaptable to diverse scenarios~\cite{gains,ying2024feature,ying2023selfoptimizingfeaturegenerationcategorical,ying2024unsupervisedgenerativefeaturetransformation,azim2024featureinteractionawareautomated,hu2024reinforcementfeaturetransformationpolymer}.
However, there are two primary limitations: 
\begin{enumerate}
    \item \textbf{Limitations 1: Permutation bias bring noise into the embedding space and lead to suboptimal performance.}
    Existing methods fail to encode the fact that feature order does not impact model performance within the learned feature subset embedding. 
    This oversight introduces bias into the embedding space, limiting the effectiveness of the search process in discovering optimal feature subsets.

    \item \textbf{Limitation 2: Convexity assumptions of the embedding space hinder effective exploration and search process.}
    Prior literature often assumes that the embedding space is convex, with the expectation that gradient-based search will locate optimal solutions. 
    In practice, this assumption rarely holds, which causes the search process to converge to suboptimal points and produce inferior feature subsets.
    
\end{enumerate}

To address these limitations, we propose \textbf{\model}, a new framework that performs \textbf{\underline{C}}ontinuous optimization for fe\textbf{\underline{A}}ture selection by integrating \textbf{\underline{P}}ermutation-invariant embeddings with a policy-guided \textbf{\underline{S}}earch strategy.
More specifically, given a large volume of feature selection records, where each record contains the indices of features within a subset and the corresponding model performance, we first design an encoder-decoder framework to capture the underlying patterns in the feature subset indices.
To achieve permutation-invariant embeddings for feature subsets, we leverage a self-attention mechanism that symmetrically computes attention scores across all input feature indices. 
This design guarantees that any permutation of the input features produces identical embeddings, ensuring that the learned representations of feature subset indices are both robust and reliable.
Considering the high computational complexity $O(N^2)$ of pairwise attention calculation, we introduce a set of inducing points to alleviate the computational burden and accelerate processing. 
These inducing points act as intermediate representations, facilitating more efficient attention computations by reducing the need for full pairwise attention. 
This approach achieves a complexity of $O(NM)$, where $M$ denotes the number of inducing points, which is substantially smaller than $N$.
After the encoder-decoder model converges, we deploy a policy-based reinforcement learning (RL) agent to explore the learned embedding space and identify the optimal feature subset.
Based on model performance, we first select the top-K feature subsets as search seeds and input them into the well-trained encoder to obtain their corresponding embeddings.
Next, we employ an RL agent to learn how to optimize these embeddings by maximizing the downstream task performance and minimizing the length of the feature subset.
Throughout this process, the exploratory nature of the RL agent helps overcome the challenges of the non-convex embedding space, enabling the search for improved embeddings.

To summarize, our contributions are three-fold:
\begin{itemize}
    \item \textbf{\textit{Problem:}}  We investigate a new framework for automated feature selection from the generative lens.
    Remarkably, we propose to solve this problem by integrating permutation-invariant embedding and policy-guided search.
    \item \textbf{\textit{Algorithm:}} We design a novel encoder-decoder architecture to embed feature selection knowledge into a permutation-invariant space, removing permutation bias. We also employ a policy-based RL agent to explore the space for optimal solutions, overcoming strong convexity assumptions.
    \item \textbf{\textit{Evaluation:}} We conduct extensive experiments on 14 real-world datasets to validate the effectiveness of our approach. The experimental results show the superior performance of \model\ over the state-of-the-art feature selection methods.
\end{itemize}

%% file: problem.tex
\section{Problem Statement}
We propose an automated feature selection framework from a generative perspective, integrating permutation-invariant embeddings with a policy-guided search strategy to enhance effectiveness.
Formally, we consider a dataset $\mathcal{D}=\{\mathbf{X},\mathbf{y}\}$, where $\mathbf{X}$ denotes features, and $\mathbf{y}$ represents the corresponding labels. 
Using existing feature selection methods~\cite{marlfs}, we can gather $p$ feature selection records, denoted by ${\{(\mathbf{f}_{i},v_{i})\}}^{p}_{i=1}$,  where each record consists of feature indices within the selected subset $\mathbf{f}_{i}$ and their corresponding model performance $v_i$.
Then, we preserve feature selection knowledge into an embedding space $\mathcal{E}$ using a permutation-invariant encoder $\omega$ and decoder $\psi$, optimized by minimizing the reconstruction loss.
Next, a policy-based reinforcement learning (RL) agent is utilized to explore the embedding space $\mathcal{E}$, aiming to identify the optimal embedding $\mathbf{E}^{*}$. 
This embedding can be used to reconstruct the optimal feature subset $\mathbf{f}^*$ by the decoder $\psi$, which maximizes downstream task performance $\mathcal{M}$.  
The optimization goal can be formulated as:
\begin{equation}
    \mathbf{f}^{*} = \psi(\mathbf{E}^{*})=\mathrm{argmax}_{\mathbf{E}\in\mathcal{E}}\mathcal{M}(X[\psi(\mathbf{E})]),
\end{equation}
Finally, $\mathbf{f}^*$ is used to select the corresponding columns of $\mathcal{D}$, forming the optimal feature space for the downstream task.

%% file: method.tex
\section{Methodology}

\subsection{Framework Overview}
\begin{figure*}[!t]
    \centering
    \includegraphics[width=\linewidth]{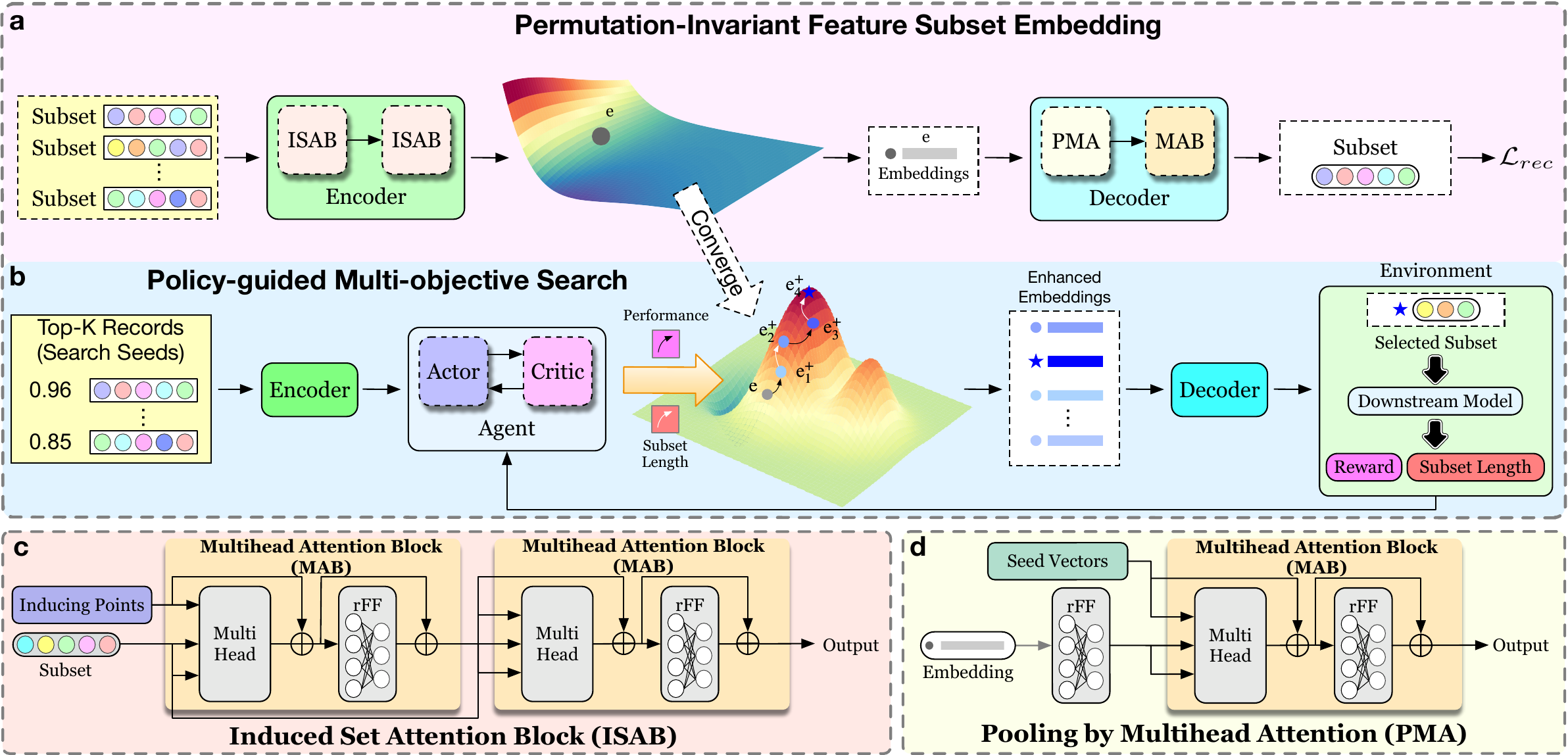}
    \captionsetup{justification=centering}
    \caption{An overview of {\model}. First, we develop an encoder-decoder structure to learn a permutation-invariant feature subset embedding space by optimizing the reconstruction loss (a).
    Then, we explore the learned embedding space through policy-guided RL search, aiming to maximize downstream task performance and minimize feature subset length (b).
    (c) and (d) illustrate the architecture of induced set attention block (ISAB) and pooling by multihead attention (PMA) respectively.}
    \vspace{-0.35cm}
    \label{framework}
\end{figure*}

Figure~\ref{framework} illustrates the overview of \model, including two main components: 1) permutation-invariant feature subset embedding (i.e., \textbf{Figure~\ref{framework} (a)}); 2) policy-guided multi-objective search (i.e., \textbf{Figure~\ref{framework} (b)}).
Specifically, given a large collection of feature selection records, each record comprises a selected feature subset and the corresponding model performance.
First, we train the permutation-invariant feature subset embedding module using these records. 
The encoder maps each feature subset to a continuous embedding by modeling pairwise relationships among feature IDs to ensure permutation invariance. 
The decoder then reconstructs the subset from the embedding to capture key feature patterns and interactions.
After the learning converges, we select the top-K selection records from the collected data and regard them as search seeds. 
Then, a policy-based RL agent explores the learned continuous embedding space. 
The updated embedding is fed into the well-trained decoder to reconstruct the feature subset. 
The search process is guided by rewards that maximize downstream task performance and minimize the feature subset length.
Finally, the feature subset achieving the highest performance is output as the optimal one.

\subsection{Permutation-Invariant Embedding Learning}
\textbf{Why permutation-invariant embedding matters.}
Feature subsets are inherently permutation-invariant, as changes in the order of selected features do not impact the downstream task performance. 
Existing works (i.e.,~\cite{gains}) often use data augmentation strategy to alleviate this burden.
In detail, they randomly shuffle the feature subset to increase the volume and diversity of the data, enhancing the embedding space.
However, with large feature subsets, this strategy may leave many cases uncovered by random shuffling.
To address this, we propose a permutation-invariant encoder-decoder.

\noindent\textbf{Training data collection.}
To learn a feature subset embedding space, we collect $p$ feature selection records, denoted by $\{(\mathbf{f}_{i}, v_{i})\}_{i=1}^{p}$, where $\mathbf{f}_{i}$  represents the feature indices of the $i$-th selected feature subset, and $v_{i}$  denotes the corresponding model performance.  
To achieve this, we use RL-based methods, specifically MARLFS~\cite{marlfs}, to automatically explore the feature space. In this method, each feature is controlled by an RL agent that decides to select or deselect it. The entire procedure is optimized to maximize downstream task performance. 
For more details, please refer to the referenced paper.

\noindent\textbf{Leveraging a Permutation-Invariant Encoder-Decoder Framework to Embed Feature Subsets.}
In the following contents, we use the feature indices of a subset $\mathbf{f}$ as an example to illustrate the calculation process within the encoder-decoder structure.

\noindent\textbf{Encoder $\omega$ :}
The encoder aims to learn a permutation-invariant mapping function $\omega$ that converts feature indices $\mathbf{f}\in\mathbb{R}^{1\times N}$ into continuous embedding $\mathbf{E}$, denoted by $\omega(\mathbf{f})=\mathbf{E}=[\mathbf{h_1},\mathbf{h_2},...,\mathbf{h_N}]\in\mathbb{R}^{N\times d}$, where $\mathbf{h_n}\in\mathbb{R}^{1\times d}$ is the $t$-th index embedding, $N$ is the number of input features, and $d$ is the hidden size of the embedding. 
Inspired by~\cite{set_tf}, we employ Multihead Attention Block (MAB) to encode pairwise interactions between features in the subset. The MAB is a modified version of the Transformer encoder block~\cite{transformer}, excluding positional encoding and dropout. Formally, the MAB with parameters $\mathbf{W}$ is defined as:
\begin{equation}
    MAB(\mathbf{Q},\mathbf{K}, \mathbf{V}) = LayerNorm(\mathbf{H}+rFF(\mathbf{H})),
\end{equation}
\begin{equation}
    \mathbf{H} = LayerNorm(\mathbf{Q}+Multihead(\mathbf{Q},\mathbf{K},\mathbf{V};\mathbf{W})),
\end{equation}  
where $rFF$ denotes row-wise feedforward layer that processes each row of feature indices independently and identically, $\mathbf{H}$ is an intermediate representation of $\mathbf{Q}$, $LayerNorm$ is layer normalization~\cite{lei2016layer}, and $Multihead$ denotes the multihead attention mechanism. Here, $\mathbf{Q}$, $\mathbf{K}$ and $\mathbf{V}$ denotes query, key and value respectively. 
In particular, we set $\mathbf{Q}=\mathbf{K}=\mathbf{V}=\mathbf{f}$ to preserve complete original information and extract pairwise relationship among all features in the input subset.
This design focuses on feature relationships rather than their order, thereby enhancing the robustness of the learned embeddings. 
However, the quadratic computational complexity $O(N^2)$ of pairwise attention becomes prohibitive when the feature number $N$ increases. 
To mitigate the computational cost, we employ a set of inducing points as intermediate representations, encoding global information about the input features. 
Specifically, in our case, each distinct feature subset and its any permuted version are mapped to a tightly clustered region in the learned embedding space. The inducing points are then trained as a small set of representative anchors in that space, effectively capturing distinct patterns from these clusters and enabling efficient global aggregation.
This strategy declines the computational complexity to $O(NM)$, where $M$ is the number of inducing points, which is significantly smaller than $N$. 
Formally, given feature indices $\mathbf{f}\in\mathbb{R}^{1\times N}$, the ISAB with $M$ inducing points $\mathbf{I}$ is defined as follows:
\begin{equation}
    ISAB_{M}(\mathbf{f}) = MAB(\mathbf{f},\mathbf{H},\mathbf{H})\in\mathbb{R}^{N\times d},
\end{equation}
\begin{equation}
    \mathbf{H} = MAB(\mathbf{I},\mathbf{f},\mathbf{f})\in\mathbb{R}^{M\times d},
\end{equation}
where the inducing points $\mathbf{I}\in\mathbb{R}^{M\times d}$ are parameterized and jointly learned with the rest of the network, and $d$ is the hidden embedding size. The ISAB comprises of two MAB layers. It is analogous to low-rank projection or autoencoder models, where the input is first compressed to low dimensional representations and then reconstructed to the original dimension. 
Specifically, we replace the query of the first MAB with the inducing points $\mathbf{I}$ so that they can attend to the entire input features, producing a low-dimensional representation $\mathbf{H}$ that encodes global information from all input features. 
Then, we replace the key and value of the second MAB with $\mathbf{H}$ to inject the global information back into each feature. Finally, to capture high-ordered feature interactions, we stack two ISAB layers to construct our encoder:
\begin{equation}
    \omega(\mathbf{f})=ISAB_{M}(ISAB_{M}(\mathbf{f}))=\mathbf{E}
\end{equation}

\noindent\textbf{Decoder $\psi$ :}
The decoder $\psi$ aims to reconstruct the feature indices $\mathbf{f}$ from the continuous embedding $\mathbf{E}$ in the learned space $\mathcal{E}$, denoted by $\psi(\mathbf{E})=\mathbf{f}$. 
To efficiently aggregate information and facilitate the reconstruction process, we integrate MAB with learnable seed vectors $\mathbf{S}\in\mathbb{R}^{K\times d}$ which is called Pooling by Multihead Attention (PMA). These seed vectors serve as prototype queries, allowing each seed vector to selectively attend to embedding in the learned embedding space. In particular, each seed vector can be seen as pooling information from a different latent aspect of the embedding, thus enhancing the efficiency of information aggregation.
Formally, given continuous embedding $\mathbf{E}\in\mathbb{R}^{N\times d}$, a PMA module with $K$ seed vectors is defined as follows:
\begin{equation}
    PMA_{K}(\mathbf{E}) = MAB(\mathbf{S},rFF(\mathbf{E}),rFF(\mathbf{E}))\in\mathbb{R}^{K\times d},
\end{equation}
where $\mathbf{E}$ denotes the input embedding to the decoder $\psi$, $rFF$ denotes row-wise feedforward layer, and $\mathbf{S}$ are parameterized and jointly learned with the rest of the network. Similar to the inducing points $\mathbf{I}$, we replace the query of the first MAB with the seed vectors $\mathbf{S}$ to ensure that they can pool information from different perspectives of $\mathbf{E}$. 
Then, to further capture the underlying interactions between $K$ outputs, we employ MAB and $rFF$ afterwards to construct our decoder:
\begin{equation}
    \mathbf{\Tilde{f}} = rFF(MAB(PMA_{K}(\mathbf{E})))\in\mathbb{R}^{1\times N},
\end{equation}
where $\mathbf{\Tilde{f}}$ is the reconstructed feature indices, and it shares the same feature index dictionary with the input feature indices $\mathbf{f}$.

\noindent\textbf{Optimization}. To effectively reconstruct the feature indices $\mathbf{f}$, we optimize the encoder and decoder by minimizing the negative log-likelihood of $\mathbf{f}$ given $\mathbf{E}$, defined as follows:
\begin{equation}
    \mathcal{L}_{rec} = - logP_{\psi}(\mathbf{f}|\mathbf{E}) = - \sum^{N}_{n = 1}logP_{\psi}(\mathbf{f}_{n}|\mathbf{h}_{n}).
\end{equation}

\subsection{Policy-Guided Multi-Objective Search}

\textbf{Why selecting search seeds matters.} As the model converges, we conduct a policy-guided RL search to identify the optimal feature subset embedding. Inspired by the initialization techniques in deep learning, we begin the search from advantageous points, referred to as search seeds, to accelerate and enhance performance. To obtain these seeds, we rank all collected records by model performance and select the top-K records to guide the search process.

\noindent\textbf{Policy-guided multi-objective search}. 
After constructing the embedding space, it must be explored to identify the optimal embedding for feature subset reconstruction. 
However, this space is not guaranteed to be convex. 
Moreover, the search for the optimal feature subset is inherently multi-objective, requiring trade-offs between maximizing downstream task performance and minimizing subset length.
Thus, to conduct effective exploration and search process without strong convexity assumptions, we employ Proximal Policy Optimization (PPO)~\cite{schulman2017proximalpolicyoptimizationalgorithms} method. 
PPO is a widely used policy gradient multi-objective method due to its stable training process and strong empirical performance. 
It maintains policy stability by clipping and trust regions, effectively balancing exploration and exploitation in high-dimensional spaces without the reliance on convexity assumptions. 
Specifically, PPO aims to learn a policy $\pi$ to guide the exploration in the learned space by maximizing a clipped surrogate objective while constraining the step size of each policy update. This approach avoids excessively large updates and thereby mitigates the risk of collapse or divergence. 
To conduct effective exploration and search process, we elaborately design the components for the PPO RL agent. Given a feature subset embedding $\mathbf{E}\in\mathbb{R}^{N\times d}$, the PPO agent takes $\mathbf{E}$ as input, and outputs the enhanced embedding $\mathbf{E^{+}}$. Then, we feed $\mathbf{E^{+}}$ into the well-trained decoder $\psi$ to reconstruct the searched feature indices $\mathbf{f^{+}}$ as state $\mathbf{s}$ for the PPO agent. We use the downstream ML task performance of the searched feature indices $\mathbf{f^{+}}$ as reward $\mathbf{R}$. Specifically, the components in our RL framework include agent, state, and reward.

\noindent\underline{$Agent.$} In our design, the PPO agent $\mathcal{P}$ manipulates the embedding $\mathbf{E}$ to generate an enhanced embedding $\mathbf{E}^{+}$, expressed as:
$
\mathcal{P}(\mathbf{E})=\mathbf{E}^{+}
$.

\noindent\underline{$State.$} The state $\mathbf{s}$ aims to describe the searched feature indices. We reconstruct the enhanced embedding $\mathbf{E}^{+}$ by utilizing the well-trained decoder $\psi$ to extract the representation of state, which can be denoted as:
\begin{equation}
    \mathbf{s} = Rep(X[\mathbf{f}^{+}]) = Rep(X[\psi(\mathbf{E}^{+})]) = Rep(X[\psi(\mathcal{P}(\mathbf{E}))]),
\end{equation}
where $Rep$ is the one-hot encoding approach, $X$ is the given dataset.

\noindent\underline{Reward.} To quantify the reward $\mathbf{R}$ obtained by the searched feature subset on the downstream ML task, we define a measurement as follows:
\begin{equation}
    \mathbf{R} = \lambda(\mathcal{M}(X[\mathbf{f}^{+}])-\mathcal{M}(X[\mathbf{f}])) + (1-\lambda)\mathcal{N}[\mathbf{f}^{+}],
    \label{eq:reward}
\end{equation}
where $\mathcal{M}$ is the downstream ML task, $\mathbf{f}$ is the original selected feature indices, $\mathbf{f}^{+}$ is the searched feature indices, $\mathcal{N}[\mathbf{f}^{+}]$ denotes the length of $\mathbf{f}^{+}$, and $\lambda$ is the trade-off hyperparameter to balance the performance and the total length of the selected feature subset.

\noindent\underline{Solving the Optimization Problem.} We train the PPO agent with multiple objectives. Specifically, we maximize a clipped surrogate objective, which is the discounted and cumulative reward during the iterative search process, while constraining the step size of each policy update. We encourage the agent to search more informative embeddings with higher model performance and fewer selected features. 
To accomplish this objective, we train the PPO agent in the actor-critic fashion. We define two separate networks for the actor and critic respectively and train them individually. We train the critic network by minimizing the difference between the cumulative discounted reward and their estimates, denoted as follows:
\begin{equation}
    \mathcal{L}_{critic} = \frac{1}{T}\sum^{T}_{t=1}(V(\mathbf{s}_t)-G_t)^{2} = \frac{1}{T}\sum^{T}_{t=1}(V(\mathbf{s}_t) - \sum^{T-t}_{k=0}\gamma^{k}R_{t+k})^{2} ,
\end{equation}
where $T$ is the total length of the trajectory, $\mathbf{s}_t$ is the state at $t$-th time step, $V(\mathbf{s}_t)$ is the estimated output of the critic network,  $G_t$ is the discounted cumulative reward at $t$-th time step, $\gamma\in [0 \sim 1]$ is the discounted factor, and $R_t$ is the ground truth reward at $t$-th time step. To train the actor network, we maximize a clipped surrogate objective while constraining the step size of each policy update, denoted as follows:
\begin{equation}
    \mathcal{L}_{actor} = \hat{\mathbb{E}}_{t}\left[ min(r_t(\theta)\hat{A}_t, clip(r_t(\theta), 1-\epsilon,1+\epsilon)\hat{A}_t)\right],
\end{equation}
where $\epsilon$ is a hyperparameter, $r_t(\theta)=\frac{\pi_\theta(\mathbf{a}_t|\mathbf{s}_t)}{\pi_{\theta_{old}}(\mathbf{a}_t|\mathbf{s}_t)}$ is the probability ratio of new policy $\pi_\theta(\mathbf{a}_t|\mathbf{s}_t)$ and old policy $\pi_{\theta_{old}}(\mathbf{a}_t|\mathbf{s}_t)$ at $t$-th time step, $\mathbf{a}_t$ is the action of the actor network at $t$-th time step, $\hat{A}_t = V(\mathbf{s}_t) - G_t$ is the estimated advantage function at $t$-th time step, and $clip(r_t(\theta),1-\epsilon,1+\epsilon)$ is the clipping function to remove the unexpected incentive for moving $r_t(\theta)$ outside of the interval [1-$\epsilon$, 1+$\epsilon$]. Eventually, by taking the minimum of the clipped and unclipped objective, we ensure that $r_t(\theta)$ is always in a trust region. Once the actor network converges, we obtain the optimal policy $\pi^*$ that chooses the most appropriate action for the exploration. In other words, it can guide the embedding $\mathbf{E}$ in the learned space $\mathcal{E}$ to shift toward the direction resulting in high downstream ML task performance without any convexity assumption.
From the training settings of aforementioned PPO agent, the well-trained agent $\mathcal{P}$ guarantees that the enhanced embedding $\mathbf{E}^{+}$ contributes to higher downstream ML task performance. Therefore, to identify the optimal feature indices, we should reconstruct the feature subset indices $\mathbf{f}^{+}$ from the enhanced embedding $\mathbf{E}^{+}$. In particular, we feed the $K$ enhanced embedding $[\mathbf{E}^{+}_{1}, \mathbf{E}^{+}_{2},...,\mathbf{E}^{+}_{K}]$ into the well-trained decoder $\psi$ to reconstruct the enhanced feature subset indices $[\mathbf{f}^{+}_{1}, \mathbf{f}^{+}_{2},...,\mathbf{f}^{+}_{K}]$. Consequently, we utilize the reconstructed feature subset indices to generate different feature subsets from the original dataset. Then, we input them into the downstream ML task to evaluate their performance and choose the feature subset indices with the highest performance as the optimal one $\mathbf{f}^{*}$.

\input{table/main_comparison}

%% file: table/main_comparison.tex
\begin{table*}[!thbp]
\centering
\vspace{-0.2cm}
\caption{Overall performance comparison. In this table, the best results are highlighted in $\bf bold$, and the second-best results are highlighted in $\underline{underlined}$. We evaluate classification (C), multi-classification (MC), and regression tasks in terms of F1-Score, Micro-F1, and 1-RAE respectively. (The higher the value is, the better the model performance is.)}
\vspace{-0.2cm}
\label{table_overall_perf}
\resizebox{\linewidth}{!}{
\setlength{\tabcolsep}{0.75mm}{
\begin{tabular}{cccccccccccccccccc}
\toprule
Dataset          & Task & \#Samples & \#Features & Original & K-Best  & mRMR & LASSO & RFE  & LASSONet &GFS &SARLFS  & MARLFS &RRA & MCDM & MEL & GAINS &\model        \\  \midrule
SpectF & C   & 267     & 44 &  75.96 &82.36&  81.41&  79.16&  78.21 &  79.16& 75.01 &75.96&  75.96& 81.23 & 79.16 & 79.69& \underline{87.38}& \textbf{89.71} \\  
SVMGuide3 & C   & 1243    & 21&  77.81 &77.02&  77.42&  74.77&  78.90&  76.51& \underline{83.70} &  78.92&  82.26  & 77.63 & 77.26 & 73.95 & 83.68& \textbf{84.22} \\  
German Credit      & C   & 1001    & 24 & 64.88 &  68.07&  66.07&  67.58 &  65.42&   66.27& 65.51 & 65.02&  68.17& 66.90 & 68.97& 64.61  &   \underline{75.34}& \textbf{77.15} \\ 
Credit Default     & C   & 30000   & 25  & 80.19 &  79.82&  80.38&  77.94&  80.22&   80.44& 80.49&  80.48&  80.47&  75.01& 74.33& 74.89 &  \underline{80.61}& \textbf{80.91} \\ 
SpamBase          & C   & 4601    & 57 & 92.68 & 91.34&  92.02&  91.69&  91.80&   89.63& 91.73 &90.94&  91.38&  90.51& 90.83& 90.47 &  \underline{92.93}& \textbf{93.13} \\  
Megawatt1  & C & 253 & 38 &  81.60 &80.08&  80.08&  83.78 &  80.08 &  82.83& 78.60 & 82.75&  78.71&  86.05& 82.83& 78.18 &   \underline{90.42}& \textbf{91.83} \\
Ionosphere        & C   & 351     & 34 & 92.85 &  92.74&  89.93&  94.14&  95.69&   94.14& 92.77& 89.82&  88.51&   92.72& 92.74& 90.16 &   \underline{97.10}& \textbf{98.55} \\  
\midrule
Mice-Protein  & MC & 1080 & 77 & 74.99 & 76.37&  78.71&  78.71&  77.29&  75.49& 75.00& 74.53&  75.01& 78.70& 78.27& 76.10 &    \underline{79.16}& \textbf{83.32} \\
Coil-20   & MC & 1440 & 400 &  96.53 & 97.58&  96.53&  94.81&  \underline{97.92} & 94.8& 92.72&   94.79&  95.14& 94.46& 93.42& 95.32 &   97.22& \textbf{98.27} \\
MNIST fashion   & MC & 10000 & 784 & 80.15 & 79.45&  80.40&  79.50&  80.70&  79.45& 80.15& 80.10&  79.90& 79.95& 79.75& 79.98 &    \underline{81.00}& \textbf{81.15} \\
UrbanSound& MC & 5702 & 16000 & 29.71&	28.74&	28.92&	27.52&	29.89&	29.18&	28.57&	30.24&	29.18&	30.50&	29.10& 24.92 &  	\underline{30.67}&	\textbf{32.52} \\
\midrule
Openml\_589          & R   & 1000    & 25 & 50.95 & 55.30&  55.28&  59.74&  55.03&  56.32& 59.74& 53.43&  53.51& 54.54& 55.43& 53.99 &     \underline{59.76}& \textbf{59.77}  \\  
Openml\_616           & R   & 500     & 50 & 15.63 & 24.86&  24.27&  29.52&  23.90&  22.98& 44.44& 25.35&  25.68& 24.67& 24.28& 42.26 &    \underline{47.39}& \textbf{48.52} \\
IQ-Dataset & R & 20000 & 38 & 98.54&	\underline{98.61}&	\underline{98.61}&	98.20&	98.58&	98.26&	90.95&	98.26&	98.27&	79.20&	98.54& 98.59 &  	98.59& \textbf{98.68}\\
\bottomrule
\end{tabular}}}
    \vspace{-0.3cm}
\end{table*}

%% file: experiment.tex
\section{Experiments}

\begin{figure*}[!h]
\centering
\subfigure[SVMGuide3]{
\includegraphics[width=4.25cm, trim = {0 0 1.5cm 0}]{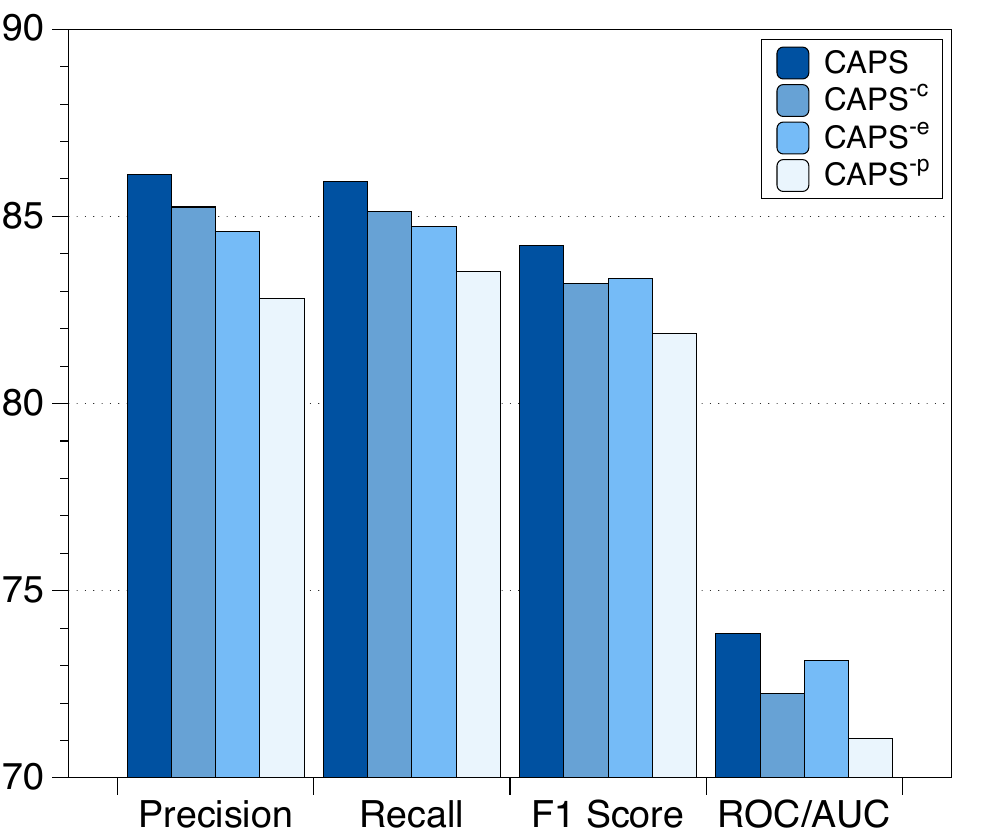}
}
\hspace{-1mm}
\subfigure[German Credit]{
\includegraphics[width=4.25cm, trim = {0 0 1.5cm 0}]{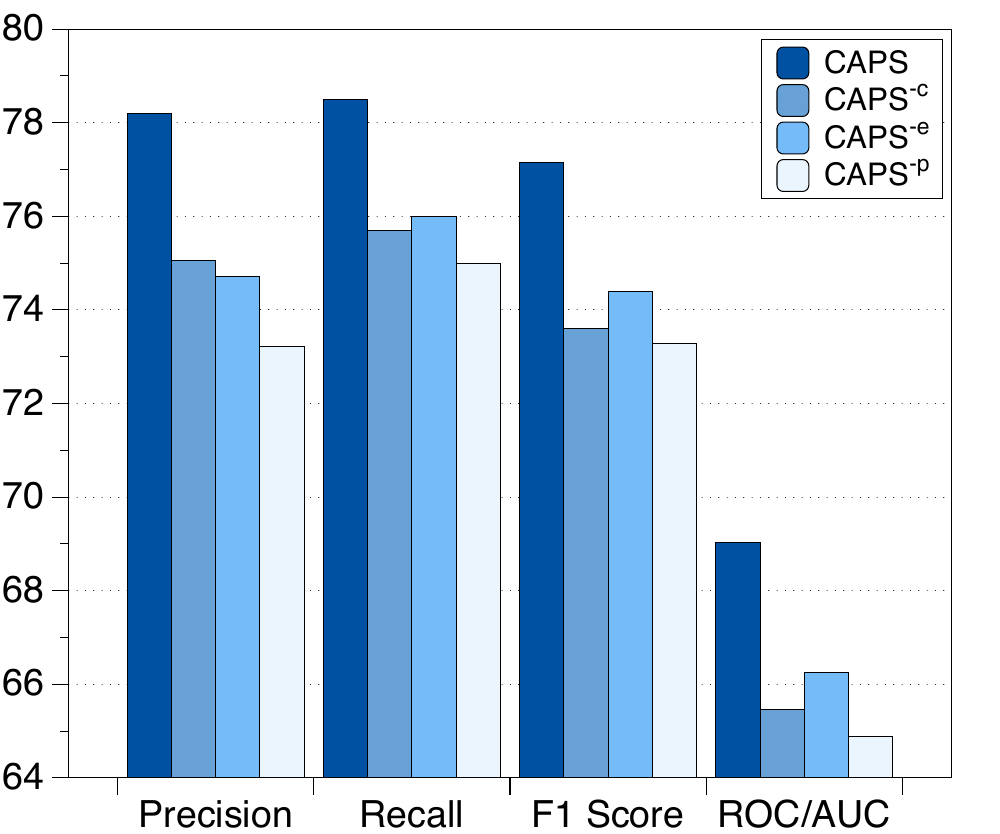}
}
\hspace{-1mm}
\subfigure[Mice Protein]{ 
\includegraphics[width=4.25cm, trim = {0 0 1.5cm 0}]{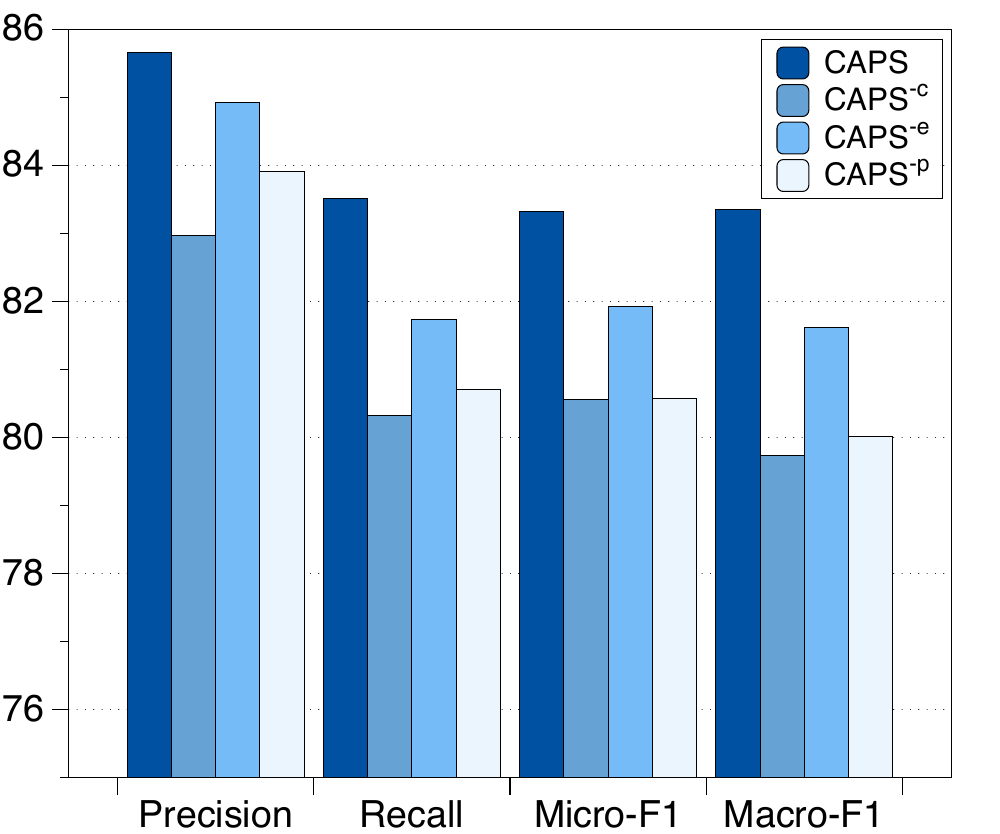}
}
\hspace{-1mm}
\subfigure[Openml\_616]{ 
\includegraphics[width=4.25cm, trim = {0 0 1.5cm 0}]{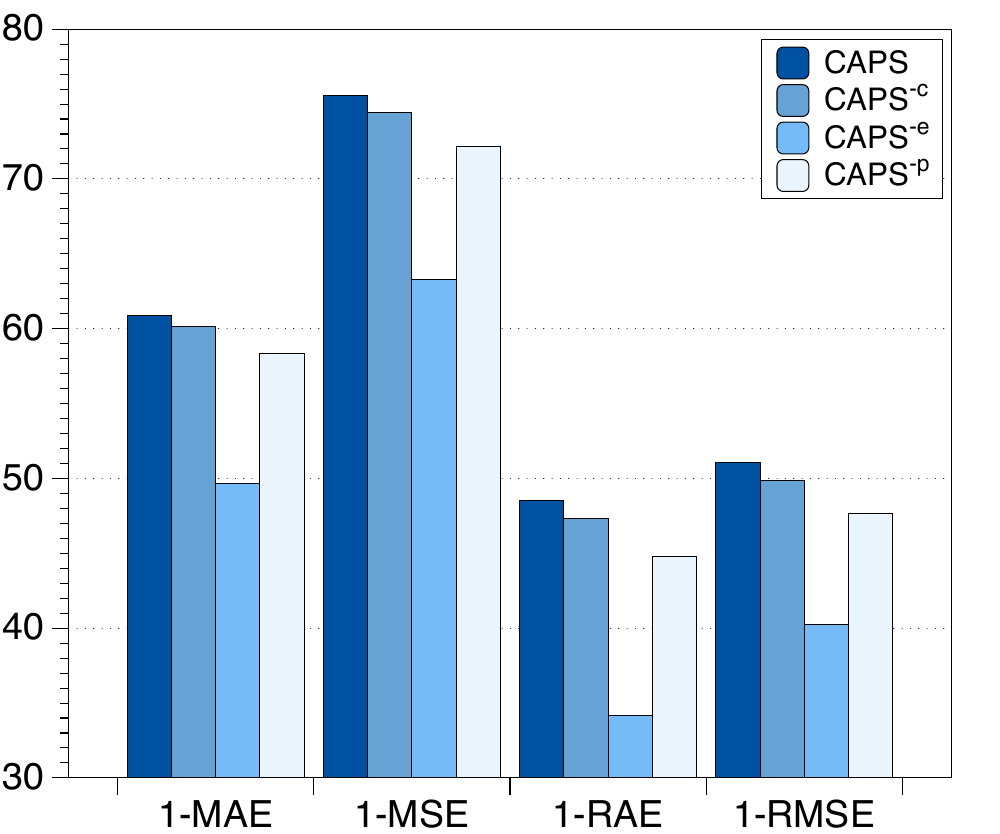}
}
\vspace{-0.4cm}
\caption{The influence of data collection (\model$^{-c}$), permutation invariance (\model$^{-e}$) and RL search (\model$^{-p}$) in \model.}
\vspace{-0.3cm}
\label{abalation_study}
\end{figure*}

\subsection{Experimental Setup}
\subsubsection{Dataset Descriptions.} 
We employ 14 publicly accessible datasets from UCI~\cite{uci}, OpenML~\cite{openml}, CAVE~\cite{cave}, Kaggle~\cite{kaggle}, LibSVM~\cite{libsvm}, UrbanSound~\cite{Salamon:UrbanSound:ACMMM:14} and OSF~\cite{osf}. These datasets can be categorized into 3 groups according to the types of ML tasks: 1) binary classification (C); 2) multi-class classification (MC); 3) regression (R). The statistics of these datasets can be found in Table~\ref{table_overall_perf}.

\subsubsection{Evaluation Metrics.}
To conduct fair comparisons and alleviate the variance of the downstream ML model, we adopt Random Forest to assess the quality of selected feature subset and report the performance of each baseline algorithm by running five-fold cross-validation method. Moreover, we conduct all experiments via the hold-out setting to further obtain a fair comparison. For binary classification tasks, we use F1-score, Precision, Recall, and ROC/AUC. For multi-classification tasks, we use Micro-F1, Precision, Recall, and Macro-F1. For regression tasks, we use 1-Mean Average Error (1-MAE), 1-Mean Square Error (1-MSE), 1-Relative Absolute Error (1-RAE), and 1-Root Mean Square Error (1-RMSE). For all metrics, the higher the value is, the better the model performance is.

\subsubsection{Baseline Algorithms.}
We compare our method with 12 widely-used feature selection methods: 
(1) \textbf{K-Best}~\cite{kbest} selects top-K features with the best feature scores; 
(2) \textbf{mRMR}~\cite{mrmr} aims to select a feature subset that comprises of the features with greatest relevance and the least redundancy to the target; 
(3) \textbf{LASSO}~\cite{lasso} conducts feature selection by regularizing model parameters, that is shrinking the coefficients of useless features into zero; 
(4) \textbf{RFE}~\cite{rfe} selects features in a recursive fashion, that is removing the weakest features until the stipulated feature numbers is reached; 
(5) \textbf{LASSONET}~\cite{lassonet} conducts feature selection by training neural networks with novel objective function;
(6) \textbf{GFS}~\cite{gfs} conducts feature selection via genetic algorithms that is recursively generating a population based on a possible feature subset first, and then using a predictive model to assess it;
(7) \textbf{SARLFS}~\cite{sarlfs} utilizes an agent to select features among all features to alleviate computational costs and views feature redundancy and downstream task performance as incentives;
(8) \textbf{MARLFS}~\cite{marlfs} is an enhanced version of SARLFS, that is building a multi-agent system to select features with the same incentive settings with SARLFS, and each agent is associated with a single feature;
(9) \textbf{RRA}~\cite{rra} is a rank integration algorithm that first gathers several selected feature subsets, and then integrates them based on the their rank in terms of statistical information;
(10) \textbf{MCDM}~\cite{mcdm} first combines the ranks of features from different feature selection algorithms to form a decision matrix, and then assigns scores to features based on this matrix for the purpose of more comprehensively and accurately selecting the most valuable features through considering multiple evaluation criteria;
(11) \textbf{GAINS}~\cite{gains} first constructs a embedding vector space by jointly optimizing an encoder-evaluator-decoder model, and then uses the gradient from well-evaluator moves the embedding point toward high performance direction to search the optimal feature subset embedding point.
(12) \textbf{MEL}~\cite{mel} divides the parent population into two sub-populations that search for optimal feature subsets independently and uses the mutual influence and knowledge sharing between the sub-populations, integrating multi-task learning with evolutionary learning. Among the introduced baseline algorithms, K-Best and mRMR are categorized as filter methods. Lasso, RFE, and LassoNet are representative of embedded methods. GFS, SARLFS, and MARLFS are considered as wrapper methods. RRA， MCDM, and GAINS are classified as hybrid feature selection methods.

\begin{figure*}[!h]
\centering
\subfigure[SpectF]{
\includegraphics[width=4.25cm, trim = {0 0 1.5cm 0}]{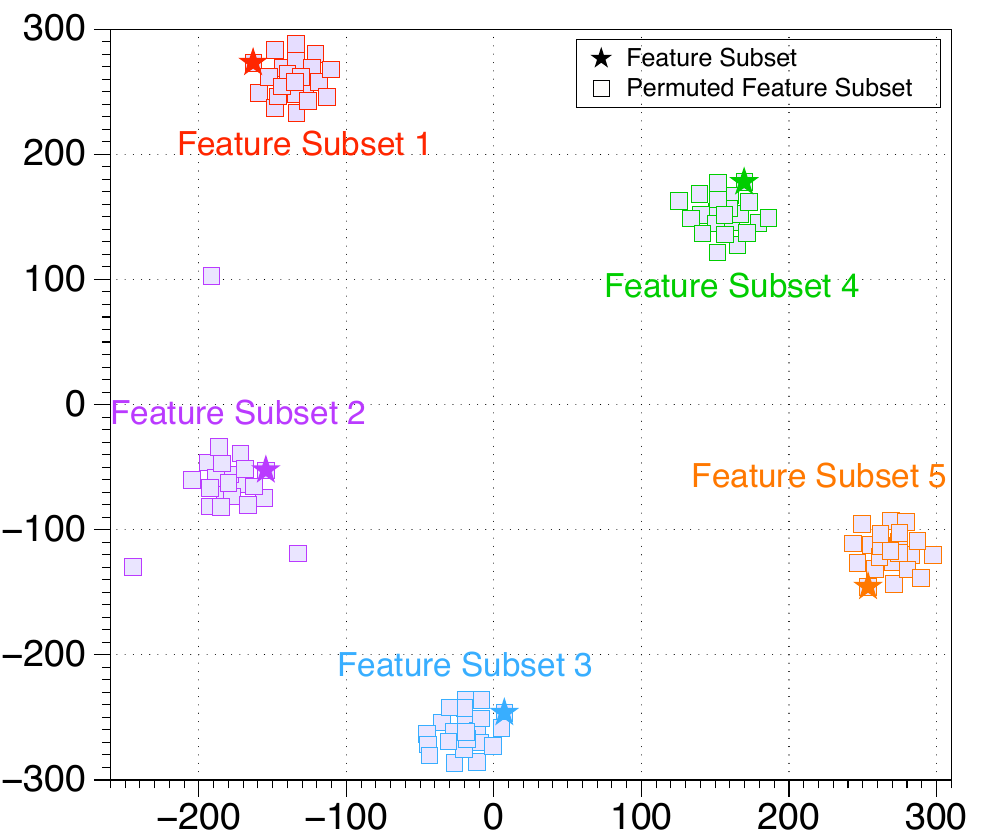}
}
\hspace{-1mm}
\subfigure[German Credit]{
\includegraphics[width=4.25cm, trim = {0 0 1.5cm 0}]{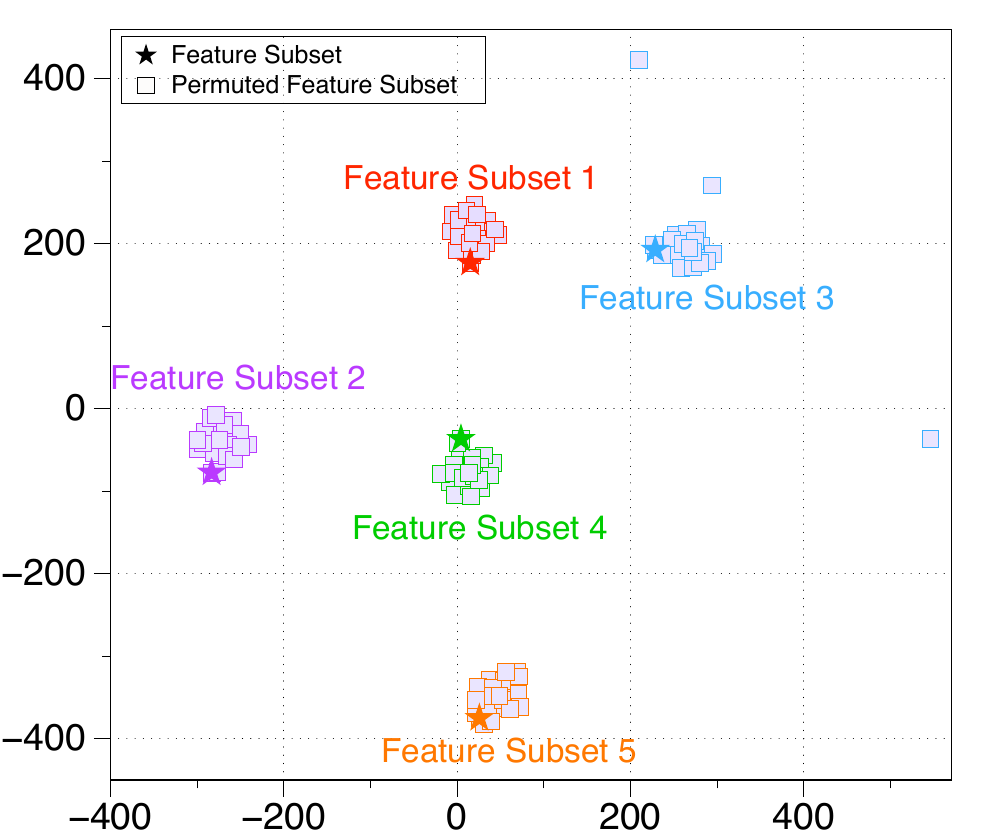}
}
\hspace{-1mm}
\subfigure[SVMGuide3]{ 
\includegraphics[width=4.25cm, trim = {0 0 1.5cm 0}]{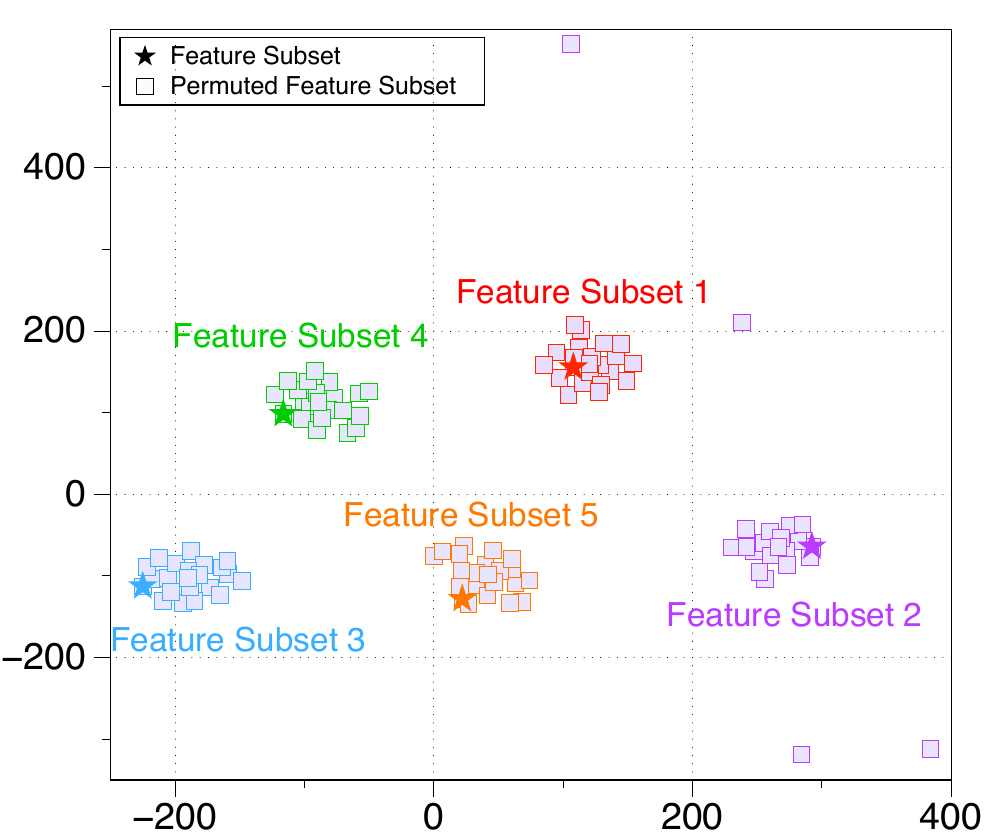}
}
\hspace{-1mm}
\subfigure[Openml\_616]{ 
\includegraphics[width=4.25cm, trim = {0 0 1.5cm 0}]{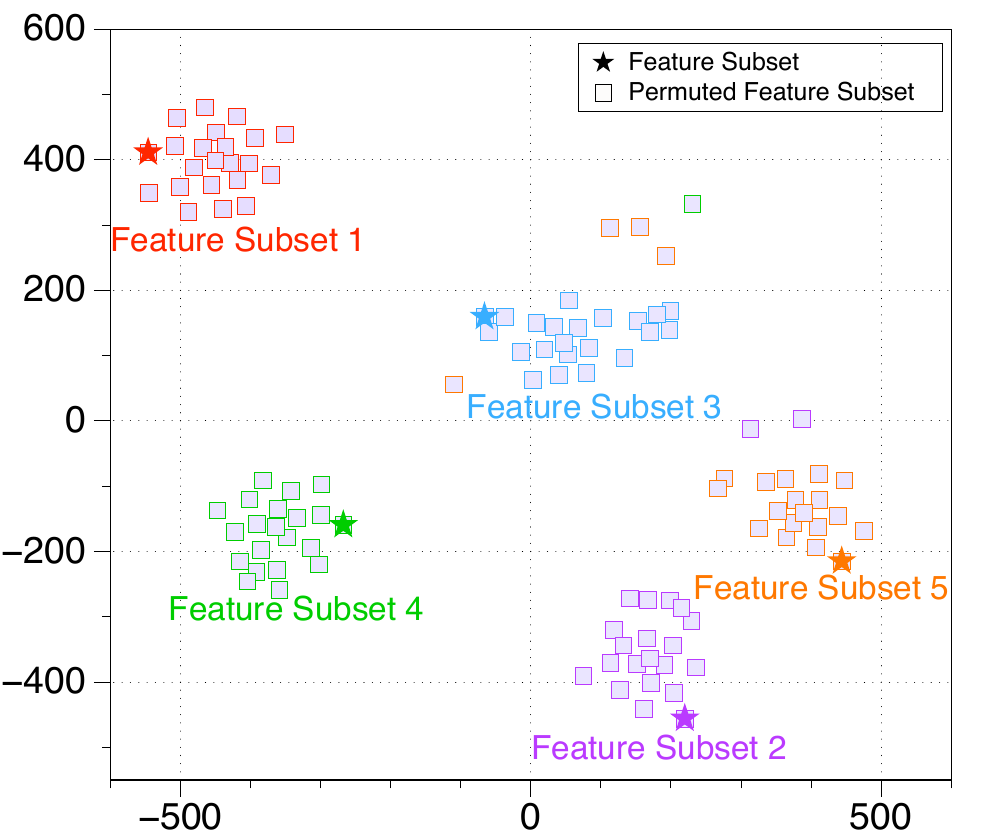}
}
\vspace{-0.4cm}
\caption{The visualization of original feature subset embeddings and permuted ones (from {\model}).}
\vspace{-0.3cm}
\label{permuted}
\end{figure*}

\subsubsection{Experimental Settings.}
We conducted all experiments on the Windows 11 operating system, AMD Ryzen 5 5600X CPU, and NVIDIA GeForce RTX 3070Ti GPU, with the framework of Python 3.10.15 and PyTorch 2.5.1.

\subsubsection{Hyperparameters and Reproducibility.}
To collect potential feature subsets and their corresponding model performance, we run MARLFS for 300 epochs. 
To increase the diversity of training data, we augment the collected feature subset-accuracy records by permuting the order of each feature subset indices 25 times. The Encoder comprises of two layers of ISAB. The Decoder comprises of 3 components, which are PMA, MAB and rFF. The number of multi-attention head is 4. The embedding size of each feature index is 128. To train the encoder and decoder, we set batch size as 64, the step size as 0.001, and the dimension of seed vectors and inducing points as 32 respectively. During the search process, we used the top 25 feature selection records as the initial search seeds to search for the optimal embedding vector. To conduct stable and robust search, we set search epoch as 10, batch size as 512, learning rate of the actor as 0.0003, learning rate of the critic as 0.001, reward trade-off as 0.1, reward discounted factor as 0.99, search step as 1000, and the clipping ratio as 0.2 respectively.

\begin{figure*}[!h]
\vspace{-0.2cm}
\centering
\subfigure[Precision]{
\includegraphics[width=4.25cm, trim = {0 0 1.5cm 0}]{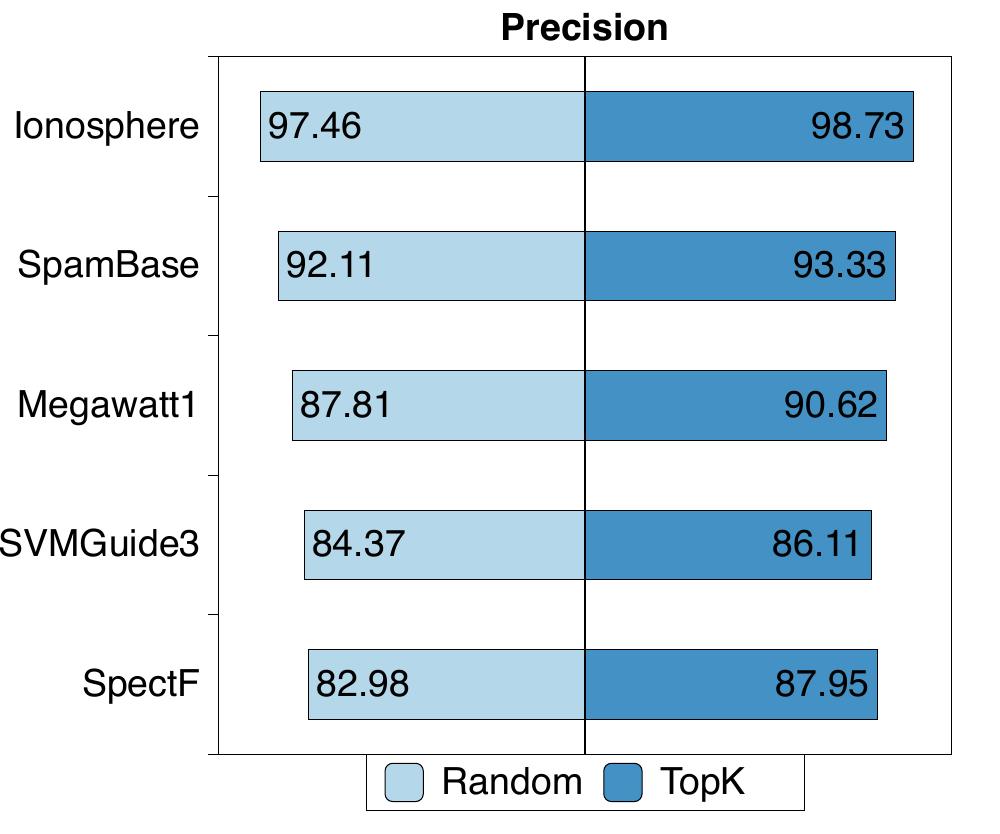}
}
\hspace{-1mm}
\subfigure[Recall]{
\includegraphics[width=4.25cm, trim = {0 0 1.5cm 0}]{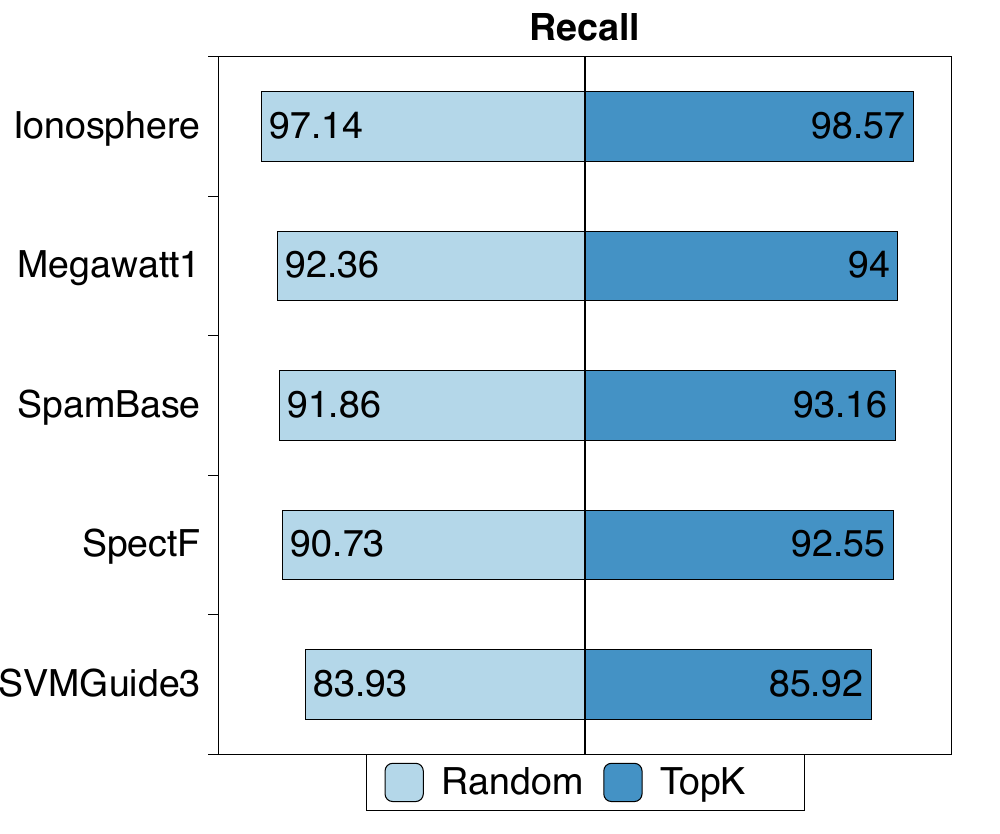}
}
\hspace{-1mm}
\subfigure[F1 Score]{ 
\includegraphics[width=4.25cm, trim = {0 0 1.5cm 0}]{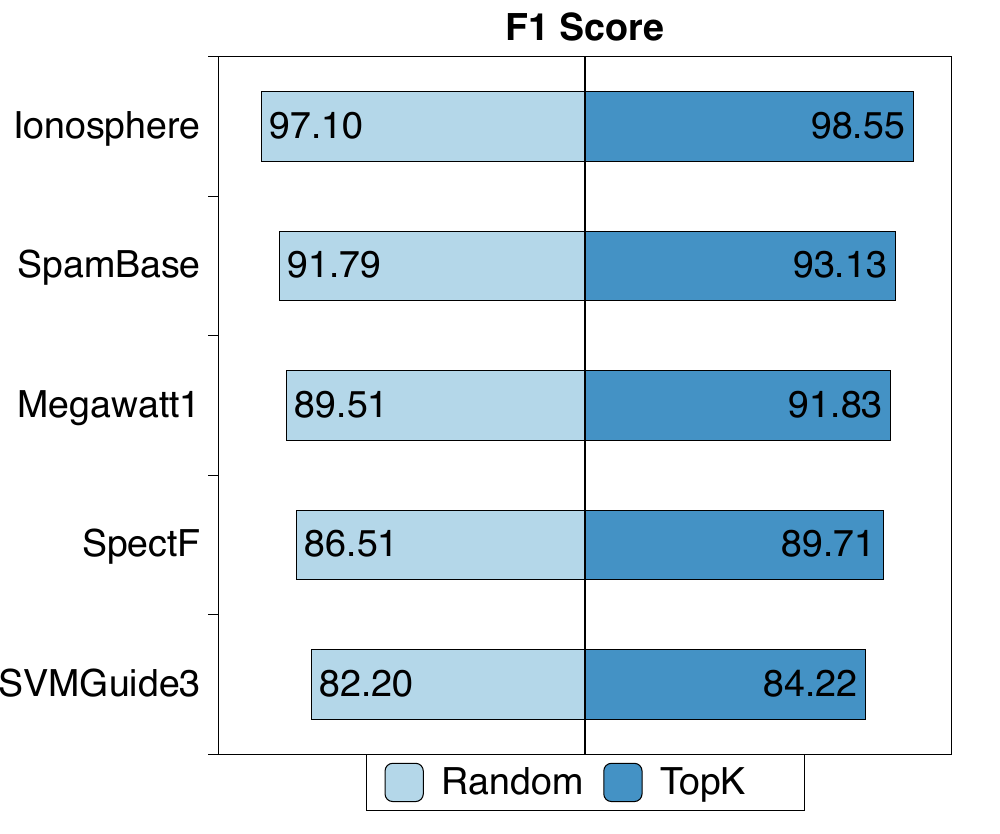}
}
\hspace{-1mm}
\subfigure[ROC/AUC]{ 
\includegraphics[width=4.25cm, trim = {0 0 1.5cm 0}]{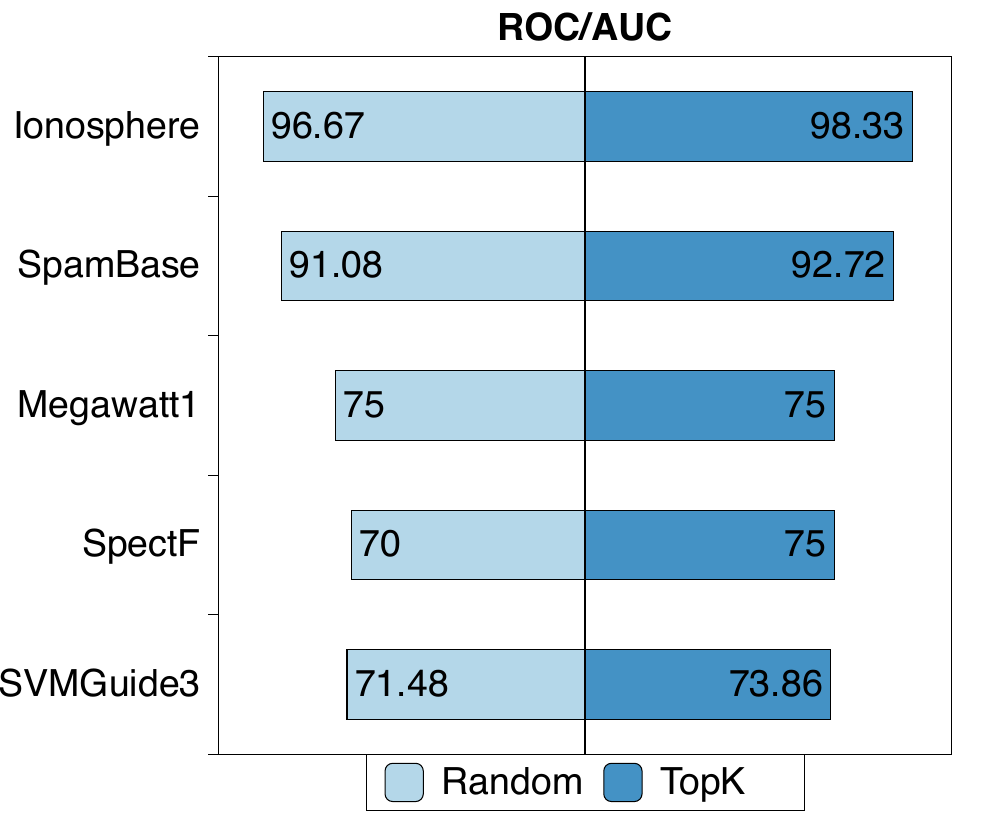}
}
\vspace{-0.2cm}
\caption{Comparison of random search seeds, and top-K search seeds in terms of Precision, Recall, F1 Score, and ROC/AUC.}
\vspace{-0.3cm}
\label{search_seeds}
\end{figure*}

\begin{figure*}[htbp]
    \centering
    \subfigure[RandomForest]{
        \includegraphics[width = 0.18\linewidth, trim = {0 0 2.5cm 0}]{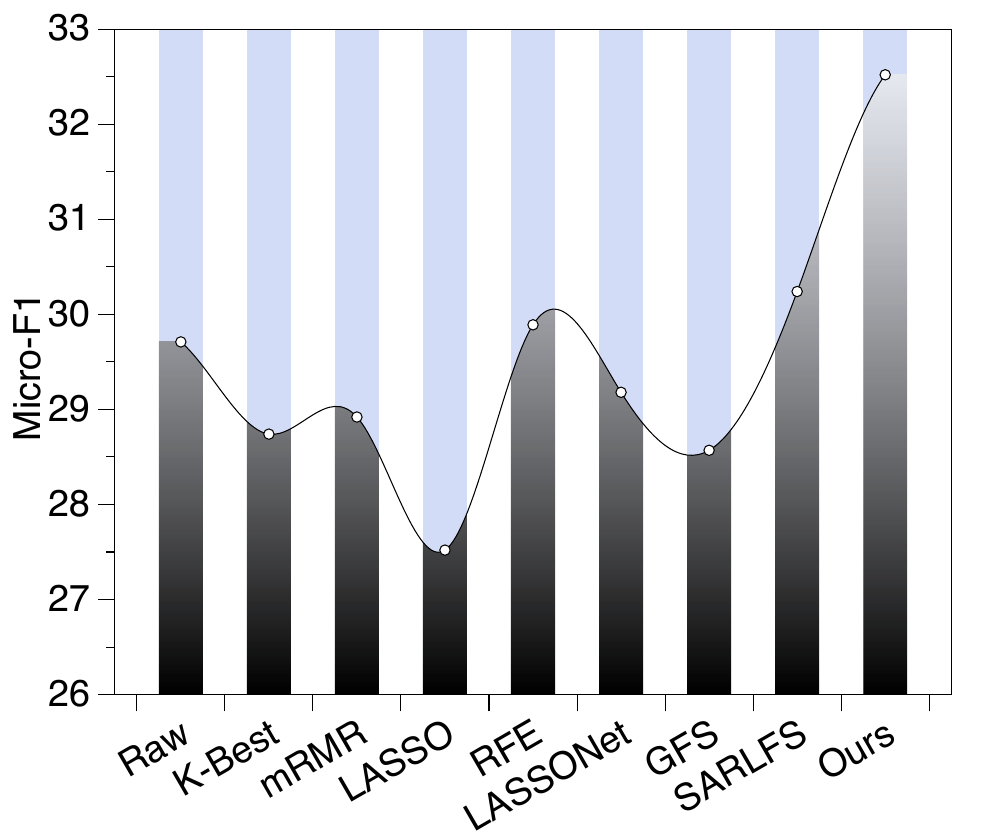}
    }
    \hfill
    \subfigure[XGBoost]{
        \includegraphics[width = 0.18\linewidth, trim = {0 0 2.5cm 0}]{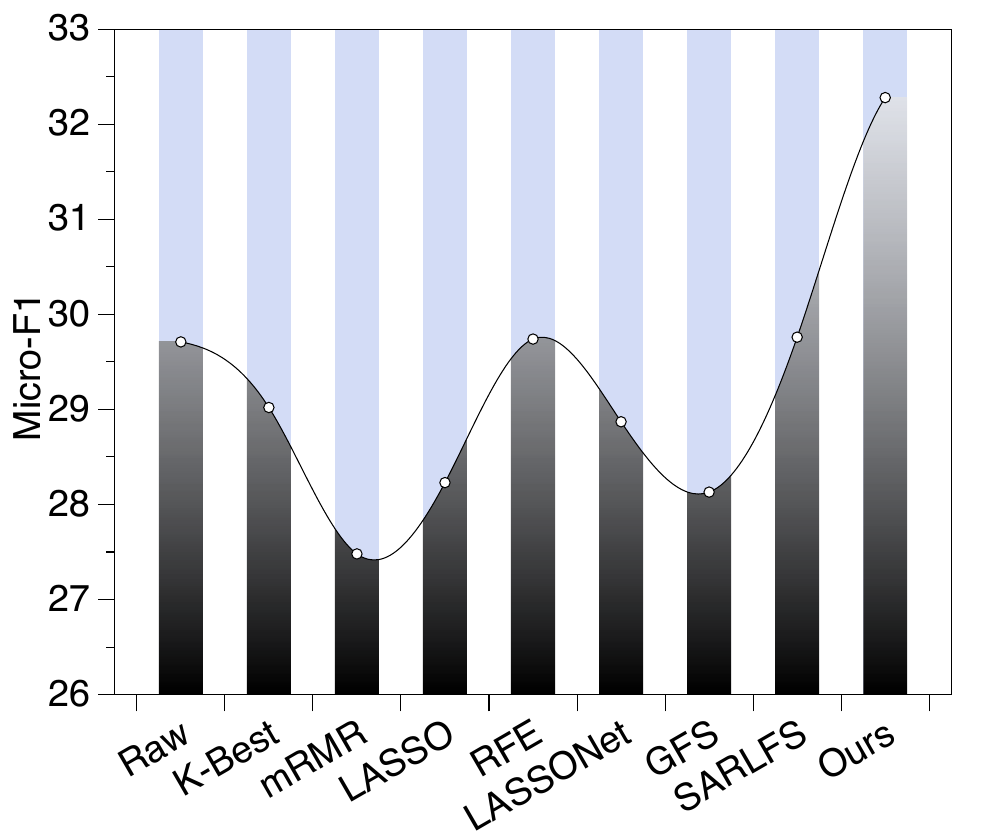}
    }
    \hfill
    \subfigure[SVM]{
        \includegraphics[width = 0.18\linewidth, trim = {0 0 2.5cm 0}]{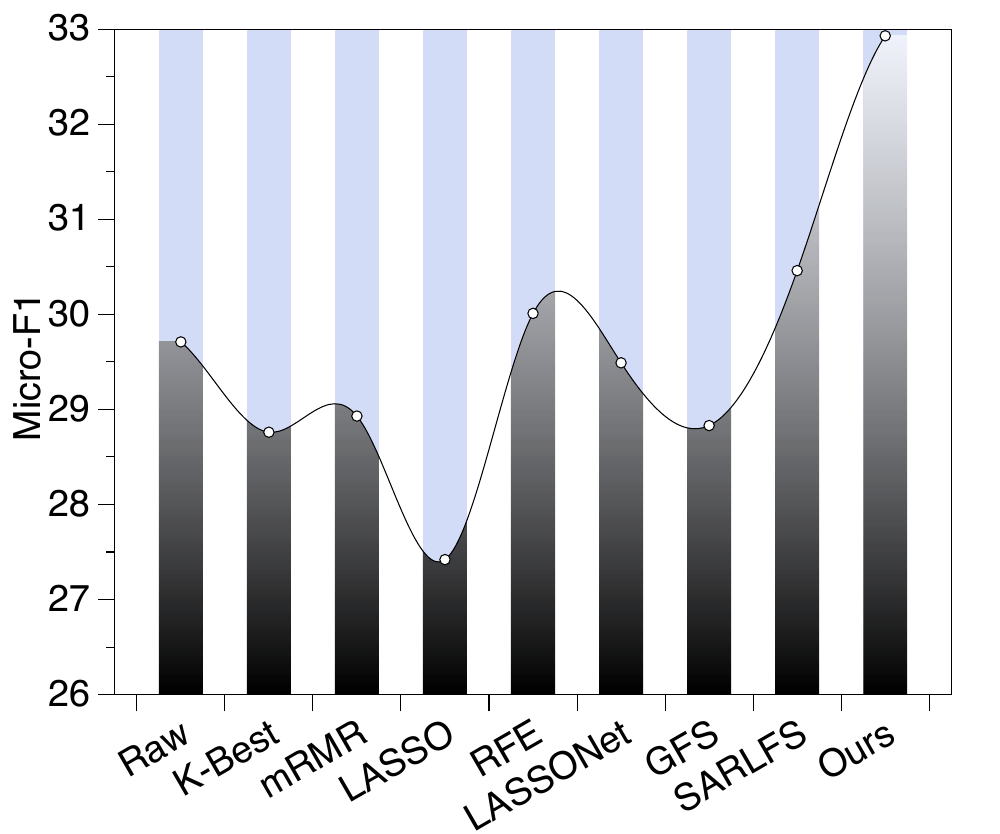}
    }
    \hfill
    \subfigure[KNN]{
        \includegraphics[width = 0.18\linewidth, trim = {0 0 2.5cm 0}]{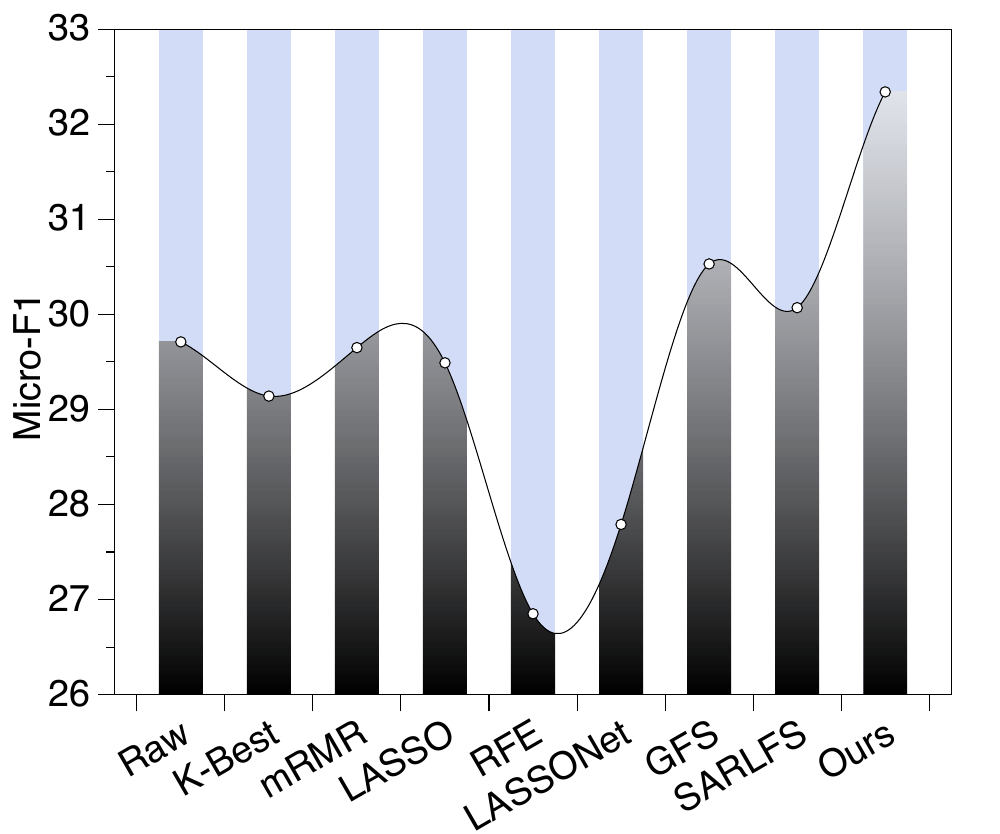}
    }
    \hfill
    \subfigure[Decision Tree]{
        \includegraphics[width = 0.18\linewidth, trim = {0 0 2.5cm 0}]{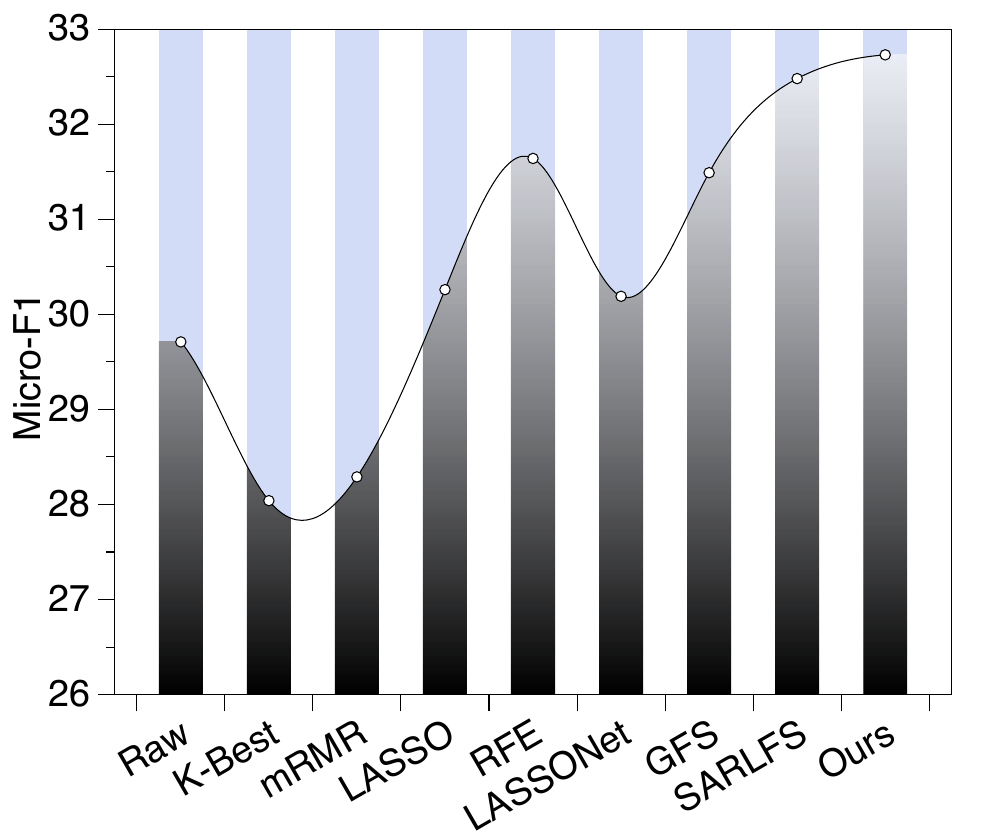}
    }
    \vspace{-0.2cm}
    \caption{Comparison of different downstream ML models in terms of Micro-F1 on UrbanSound.}
    \vspace{-0.3cm}
    \label{table_robust}
\end{figure*}

\subsection{Experimental Results}
\subsubsection{Overall performance.}
This experiment is conducted to assess the overall performance of our method. We compare the performance of our method and baseline algorithms. Table~\ref{table_overall_perf} shows the comparison of our method and 12 baseline algorithms in terms of F1-score, Micro-F1, and 1-RAE. The results demonstrate that our method outperforms other baseline algorithms across all domains and various tasks. 
There are two possible reasons for this observation: 
1) The permutation-invariant encoder-decoder framework preserves feature subset knowledge in the learned embedding space, focusing on feature pairwise interactions rather than their order and removing the permutation bias in the embedding space;
2) The policy-based RL agent effectively explores the learned continuous embedding space and identifies the superior feature subset embeddings, overcoming the challenges of the non-convex space and converging to the global optimal feature subset embedding.

\subsubsection{Ablation Study.}
We conduct this experiment to study the impact of the data collection, permutation-invariant embedding, and policy-guided search on the model performance. 
We implement three model variants: 
1) \model$^{-c}$ collects the feature subset-accuracy pairs randomly instead of employing basic and RL-based data collectors; 
2) \model$^{-e}$ utilizes sequential encoder-decoder model to learn the continuous embedding space.
3) \model$^{-p}$ replaces the policy-guided search with Genetic Algorithm (GA) to explore the embedding space and identify the optimal feature subset embedding.
We randomly choose four datasets to conduct this experiments, which are SVMGuide3, German Credit, Mice Protein, and Openml\_616. 
For each type of task, we evaluate the model performance from four different perspectives. For binary classification tasks, we use Precision, Recall, F1-score, and ROC/AUC. 
For multi-classification tasks, we use Precision, Recall, Micro-F1, and Macro-F1. 
For regression tasks, we use 1-MAE, 1-MSE, 1-RAE, and 1-RMSE. 
Figure~\ref{abalation_study} shows the overall experiment results. 
We observe that the performance of \model\ is higher than \model$^{-c}$, \model$^{-e}$ and \model$^{-p}$ on all datasets. 
The potential reasons for this observation are: 
1) the feature selection records generated by the data collectors are more accurate and robust, constructing a more comprehensive and effective continuous embedding space for searching the optimal feature subset; 
2) sequential encoder-decoder model fails to embed permutation invariance in the embedding space, introducing permutation bias and leading to local optima; 
3) policy-guided RL agent conducts effective exploration for optimal feature subset, eliminating the reliance on convexity assumptions and avoiding convergence to local optima.
Thus, this experiment demonstrates the significance of data collector, permutation-invariant embedding and policy-guided search for enhancing the performance of \model.

\subsubsection{Study of the permutation sensitivity of learned feature subset embeddings.} We conduct this experiment to study the permutation sensitivity of learned feature subset embeddings. We randomly choose four datasets and visualize the embeddings of the original feature subsets along with their corresponding permuted versions. In detail, we first randomly collect five distinct feature subset as the original representatives. Each feature subset is subjected to a permutation operation for 20 times to generate the corresponding permuted representatives. 
Then, we employ the well-trained encoder to obtain the embeddings for both the original and permuted feature subsets. Subsequently, we use T-SNE to map these embeddings into a 2-dimensional space for visualization. 
Figure~\ref{permuted} shows the visualization results, in which each color represents a unique feature subset along with its corresponding permuted versions. The embeddings of the original feature subset and the corresponding permuted versions are represented by pentagrams and squares respectively. 
We find that the distribution locations of the permuted feature subset embeddings are consistently clustered around their corresponding original feature subset embedding. 
A potential reason for this observation is that the learned continuous embedding space inherently captures permutation invariance, ensuring that variations in feature order do not impact the encoded embedding distribution. Therefore, this experiment demonstrates that our proposed encoder-decoder embeds 
feature subset knowledge into a permutation-invariant continuous embedding space, successfully removing the permutation bias.

\subsubsection{Study of the impact of search seeds.}
We conduct this experiment to study the impact of initial search seeds for the RL-based search process. 
To mitigate bias, we conduct the comparison on eight randomly chosen datasets and in term of four metrics, which are Precision, Recall, F1 Score, and ROC/AUC. 
We compare the search process using random K historical records with the one using top-K historical records based on model performance. 
Figure~\ref{search_seeds} shows the overall comparison results. 
The experimental results demonstrate that the search process with top-K records as initial search seeds consistently outperforms the one without across all metrics. 
A plausible explanation for this observation is that the starting of the search process is sensitive to the initial starting points. Initializing the search with top-K historical records not only provides a more promising starting region in the embedding space but also reduces the likelihood of exploring less relevant or suboptimal areas. The RL-based search agent can leverage the historical information to form a lower bound and focus on refining already competitive solutions. This leads to faster convergence and more stable performance across various datasets. Another noteworthy observation is that using top-K seeds helps to alleviate the risk of becoming trapped in local optima since these seeds often span multiple good regions in the embedding space. 
In contrast, random seeds might guide the search process toward less advantageous regions, increasing the variance in final performance. 
Therefore, these observations demonstrate that search seeds are crucial for the RL agent to explore the space and identify the optimal feature subset akin to the significance of initialization for deep neural networks.

\subsubsection{Study of the robustness of \model\ over downstream ML tasks.}
We conduct this experiment to assess the robustness of \model\ by changing the downstream ML model to Random Forest (RF), XGBoost (XGB), Support Vector Machine (SVM), K-Nearest Neighborhood (KNN), and Decision Tree (DT) respectively. 
Figure~\ref{table_robust} shows the comparison results on UrbanSound dataset in terms of Micro-F1. 
The results demonstrate that \model\ outperforms other baselines regardless of downstream ML models. 
A potential reason for this observation is that the data collector can customize the feature selection records based on the downstream ML task. 
Then, the encoder-decoder model is capable of embedding the preference and properties of the ML model into the continuous embedding space. 
Finally, by leveraging these information, the RL agent can reach the global optimal embedding point instead of local optima. 
Thus, this experiment successfully demonstrates the robustness of \model.

\begin{figure}[!ht]
    \centering
    \includegraphics[width=\linewidth]{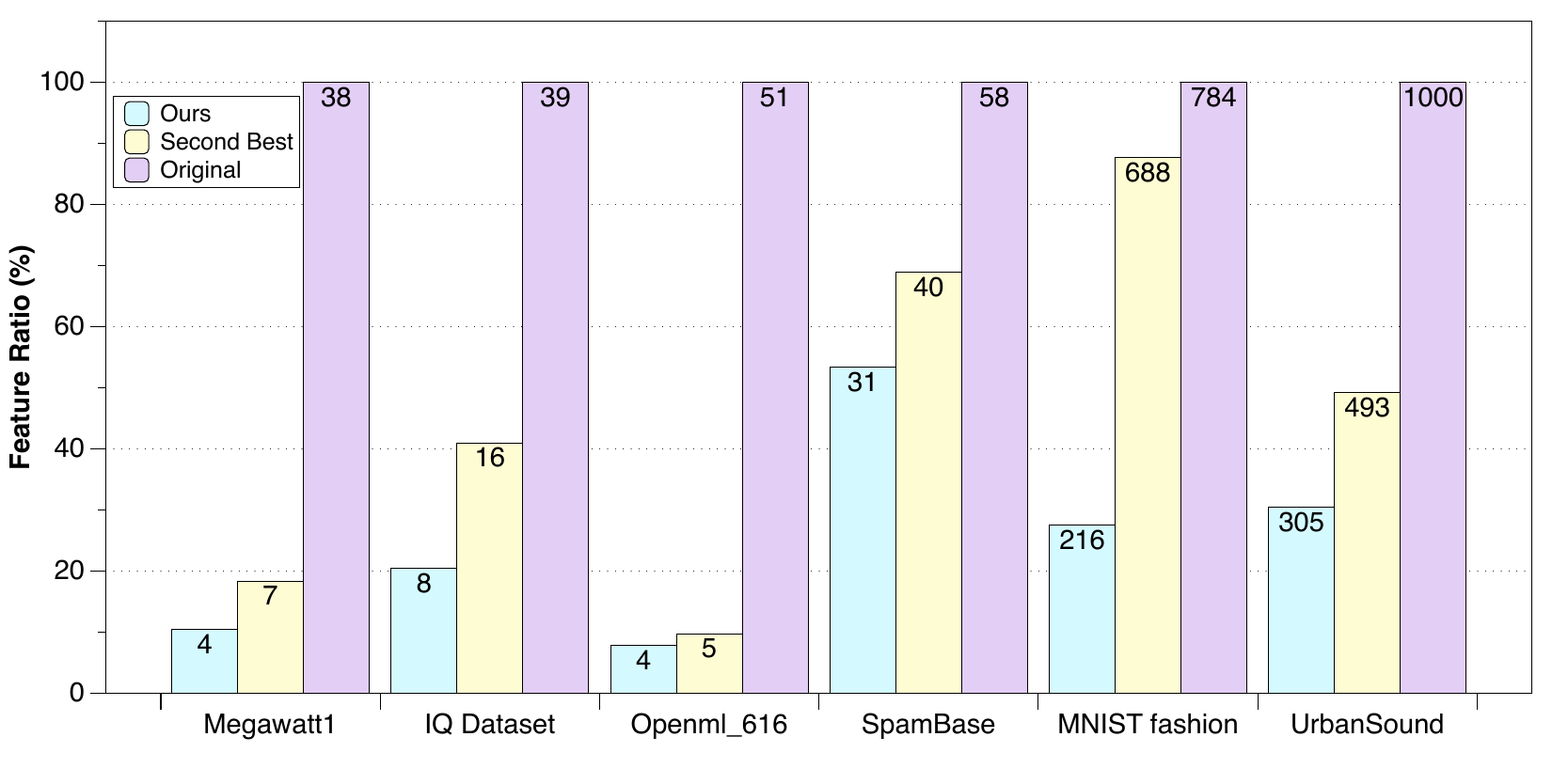}
    \caption{Comparison of the selected feature subset size of our method,  the best baseline algorithm (Second Best), and the original feature set (Original).}
    \label{feat_num}
\end{figure}

\begin{figure}[!thbp]
        \centering
        \subfigure[Clipping Trade-off]{
        \includegraphics[width=0.47\linewidth,trim = {2cm 0 2cm 0}]{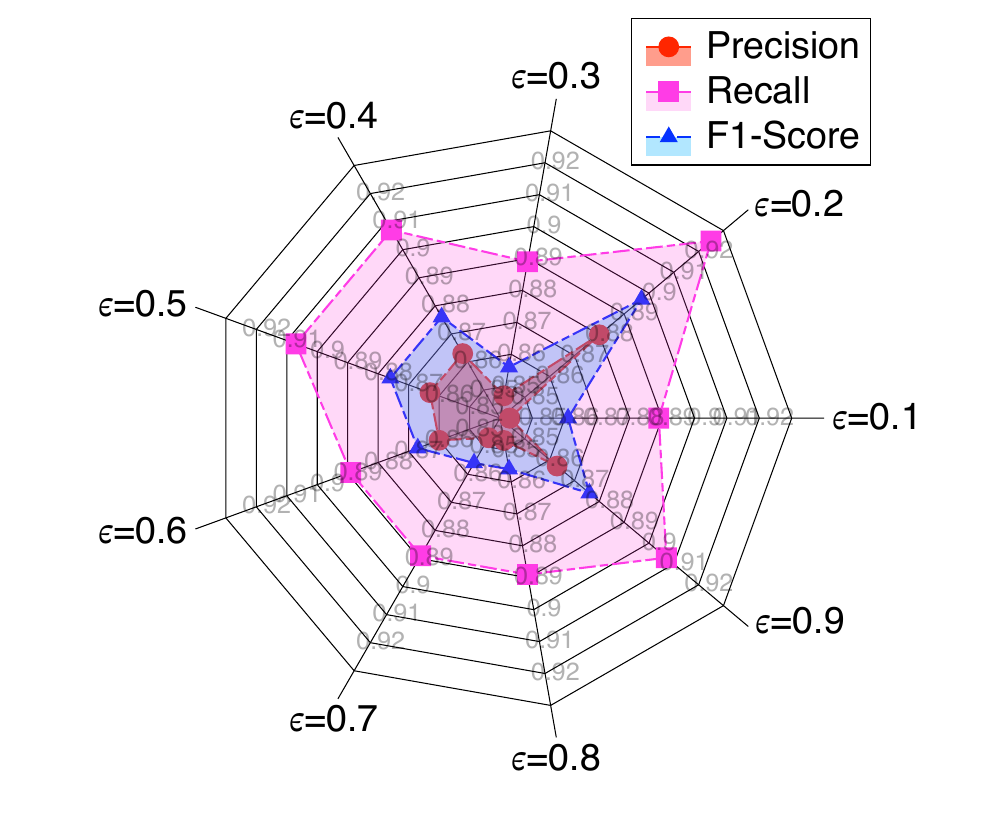}
        }
        \hfill
        \subfigure[Reward Trade-off]{
        \includegraphics[width=0.47\linewidth,trim = {2cm 0 2cm 0}]{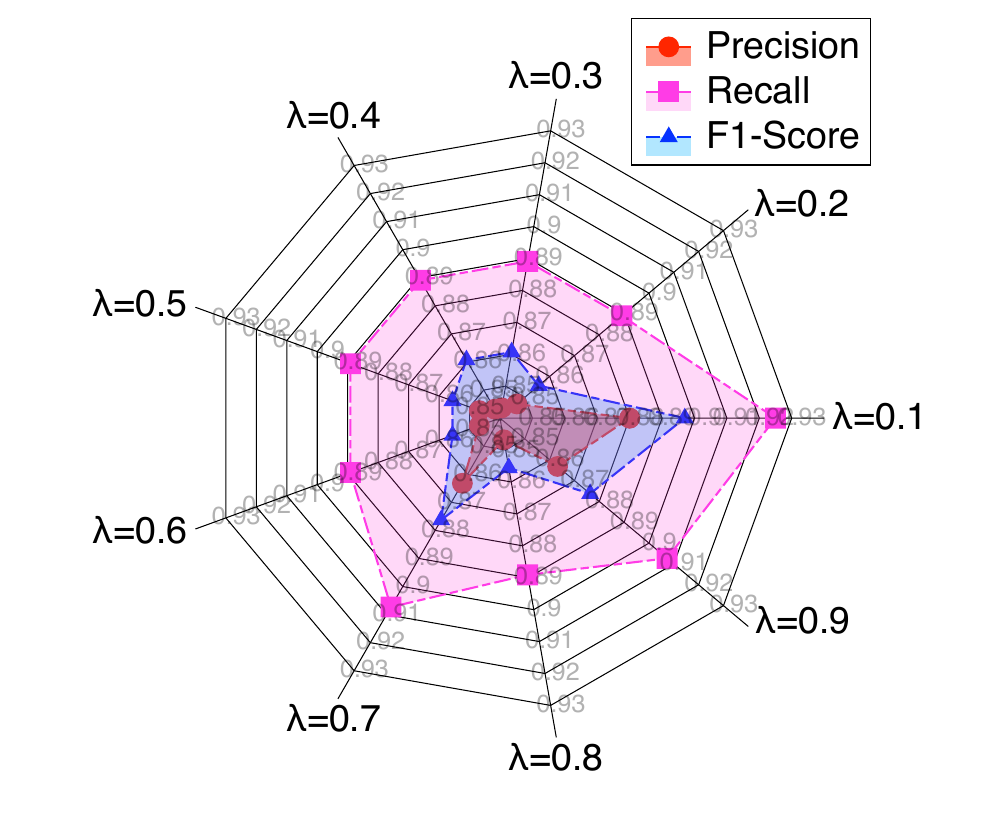}
        }
        \caption{The hyperparameter sensitivity test on SpectF.}
        \label{trade}
\end{figure}

\subsubsection{Study of the size of selected feature subsets.}
We conduct this experiment to assess the ability of the RL agent to optimize multiple objectives simultaneously, improving model performance while reducing selected features.
We conduct the comparison on six randomly selected datasets, and compare the size of selected feature subset of our method with the best-performing baseline algorithm and the original feature set. 
Figure~\ref{feat_num} demonstrates the results of overall comparison in terms of the ratio of selected feature subset size to original feature set size. 
The ratio indicates the proportion of selected features relative to the original features. 
The results show that the selections of our method are consistently smaller than the ones selected by the second-best baseline algorithms on all the datasets. 
Although our selected feature subsets contain noticeably fewer features, they consistently achieve superior or comparable model performance. 
A potential reason for this observation is that the policy-guided RL agent is capable of optimizing the learned embeddings by maximizing the downstream task performance and minimizing the length of the selected subset. 
Therefore, this experiment demonstrates that the policy-guided RL agent excels at striking a balance between model performance and feature-efficiency.

\subsubsection{Study of the Hyperparameter Sensitivity of Search Process.}
We conduct this experiment to study the sensitivity of the search process performance to the the clipping trade-off $\epsilon$ and the reward trade-off $\lambda$.
As part of the surrogate objective, the clipping trade-off $\epsilon$ can stabilize training and prevent excessively large updates to the original policy. 
Small value of $\epsilon$ results in stricter policy updates, ensuring that each step is small and stable, but at the risk of limiting exploration and making it harder for the policy to improve. 
Conversely, larger value of  $\epsilon$ results in more generative policy updates, facilitating quicker exploration of the policy space.
In other words, a smaller $\epsilon$ prioritizes exploitation over exploration, and vice versa.
The reward trade-off $\lambda$ is designed for the RL agent to balance the attention on the feature subset performance and the total length of the subset from equation~\ref{eq:reward}. 
In particular, a larger $\lambda$ prioritizes the feature subset performance over total length, and vice versa. 
We set both $\epsilon$ and $\lambda$ vary from 0.1 to 0.9, then train the \model\ on Spectf. The model performance is shown in Figure~\ref{trade} in terms of Precision, Recall and F1-Score. Overall, we observed that \model\ achieves best performance when $\epsilon$ is set to 0.2 and $\lambda$ is set to 0.1. From an empirical perspective, PPO has been shown to perform robustly across a wide range of tasks, with $\epsilon$ typically set to 0.2 in many standard tasks~\cite{schulman2017proximalpolicyoptimizationalgorithms}, which aligns with our findings. 
As for the reward trade-off $\lambda$, the model performance remains relatively stable across most values of $\lambda$ and occasionally increases when $\lambda$ is set to 0.1. 
These findings provide insights into how $\epsilon$ and $\lambda$ influence the search results and how we should choose optimal values for them.

\subsubsection{Case study.}
We conduct this experiment to examine the traceability of {\model}. 
We rank the top 7 significant features for prediction in both the original feature set and the selected feature subset using the IQ-Dataset.
Figure~\ref{case_study} (a-b) show the top 7 significant features among 38 original features and the selected features of {\model} respectively. 
The predictive label is the summation of NVER\_SC\_119 and VERB\_SC\_119, which is the summed standardized score from 1 to 19 for the Nonverbal Intelligence scale and the summed standardized score from 1 to 19 for the Verbal Intelligence scale respectively. Figure~\ref{case_study} (a) illustrates that these two crucial features are absent from the top 7 features among the original 38 features. 
However, from Figure~\ref{case_study} (b), we observe that {\model} successfully selected these two features from 38 original features. 
A potential reason for this observation is that {\model} is capable of capturing the complicated interactions among features as well as the causality between features and predictive label. 
Therefore, this experiment demonstrates the traceability of the feature subset selected by {\model}, highlighting its effectiveness from a different perspective.

%% file: related.tex
\section{Related Works}
\textbf{Feature selection} screens discriminative features from high dimensional data, improving model efficiency and performance by reducing dimensionality and shortening training time.
Conventional feature selection methods are categorized into three main types: 
1) Filter methods evaluate feature importance using statistical metrics (e.g., correlation, mutual information and chi-square tests)~\cite{kbest,hall1999feature,forman2003extensive,yu2003feature,mrmr,biesiada2008feature,ding2014identification}.
However, such methods generally ignore the dependency and interactions among features.
2) Wrapper methods formulate feature selection as a search problem, training a machine learning (ML) model as the evaluation criterion and selecting the subset that yields the best performance~\cite{liu2021efficient,kim2000feature,narendra1977branch,kohavi1997wrappers,gfs}. However, such methods require excessive model training, resulting in high computational costs. 
3) Embedded methods integrate feature selection into the model training process as a part of the optimization objective to evaluate the importance of features~\cite{lasso,sugumaran2007feature,lassonet,rfe,kumagai2022few,koyama2022effective}. However, such methods heavily rely on the underlying model structure. 
Recently, generative methods have attracted substantial attention due to the success of generative AI~\cite{gains,ying2024feature,grfg,2022traceable,2023traceable,ref_arxiv,fedcaps, urbanplanning_rui, catf_rui}.
For instance,~\cite{gains} introduces a framework that encompasses  a sequential encoder, an accuracy evaluator, a sequential decoder, and a gradient ascent optimizer, embedding discrete feature selection knowledge into a continuous embedding space and conducting gradient-ascent search process for the optimal feature subset.
However, this approach is limited by: 
1) incapable of encoding the permutation invariance into the embedding space; 
2) the reliance on the convexity assumptions about the feature subset embedding space.
To address these drawbacks, we integrate a permutation-invariant embedding mechanism with a policy-guided search strategy in {\model}. 
The permutation-invariant embedding is designed to mitigate the sensitivity to feature ordering, constructing a more robust embedding space that preserves feature selection knowledge. 
The policy-guided search leverages a reinforcement learning strategy to effectively explore this space, identifying better feature subsets without relying on strong convexity assumptions.

\begin{figure}[!t]
        \centering
        \subfigure[Original Feature Set]{
        \includegraphics[width=0.47\linewidth,trim=0.5cm 0.3cm 0.5cm 0.3cm, clip]{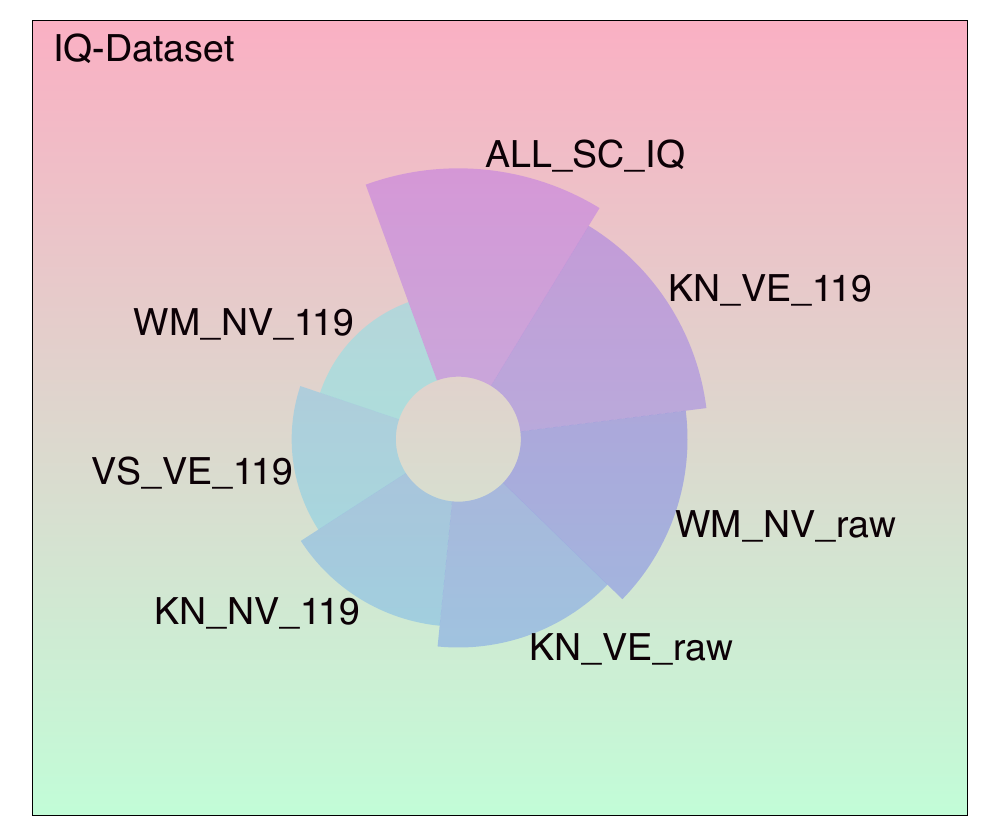}
        }
        \hfill
        \subfigure[{\model}-selected Feature Subset]{
        \includegraphics[width=0.47\linewidth,trim=0.5cm 0.3cm 0.5cm 0.3cm, clip]{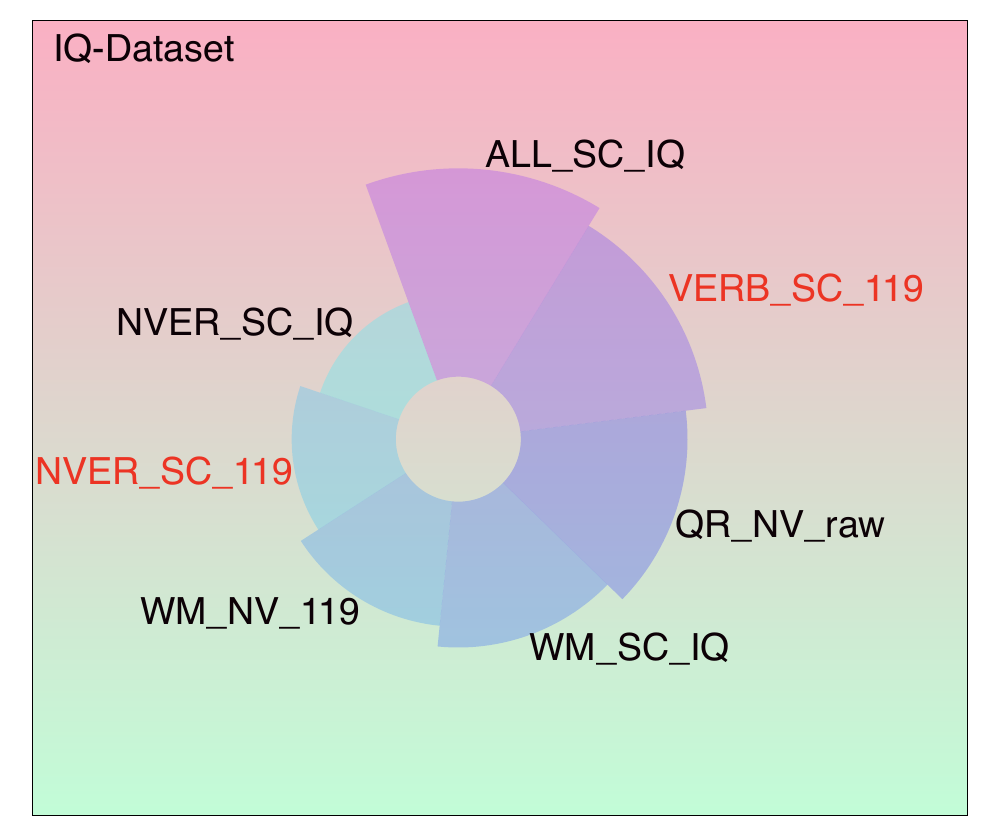}
        }
        \caption{Case Study. Comparison of traceability on the original feature set and selected feature subset.}
        \label{case_study}
\end{figure}

%% file: conclusion.tex
\section{Conclusions}
In this paper, we propose a generative automated feature selection framework that integrates permutation-invariant embeddings and policy-guided search to overcome permutation bias and convexity assumptions in existing methods.
In detail, we first train a permutation-invariant feature subset embedding module using feature selection records by optimizing the reconstruction loss. The encoder maps feature subsets to continuous embeddings, while the decoder reconstructs the subsets from these embeddings.
Moreover, considering the high computational cost of the pairwise interaction calculations, we leverage a set of inducing points as intermediate representations. 
These inducing points are expected to extract information about the input feature subset, contributing to capture global feature pattern and facilitate more efficient attention calculations by reducing the need for full pairwise attention. 
As the model converges, we employ a policy-guided RL agent to explore the learned space for superior feature subsets, overcoming the reliance on convex assumptions and mitigating the risk of converging to local optima.
Finally, extensive experiments demonstrate that our framework offers significant insights: 
1) Feature order sensitivity introduces permutation bias into the embedding space and undermines the embedding search process.
2) Policy-guided RL search prioritizes high-potential regions and eliminates the reliance on convexity assumptions, avoiding convergence to local optima.
These insights emphasize the potential of reducing permutation sensitivity and overcoming non-convexity to improve the effectiveness and efficiency of automated feature selection. 
In the future, a promising direction will be to enhance the efficiency of the embedding search process. This can be achieved by reducing its strong reliance on a well-trained decoder for subset reconstruction.



%% file: sample-base.bib
@String{Computing = "Computing" }

@String{Computer = "{IEEE} Computer" }

@String{Springer = "Springer-Verlag" }

@InProceedings{set_tf,
  title = 	 {Set Transformer: A Framework for Attention-based Permutation-Invariant Neural Networks},
  author =       {Lee, Juho and Lee, Yoonho and Kim, Jungtaek and Kosiorek, Adam and Choi, Seungjin and Teh, Yee Whye},
  booktitle = 	 {Proceedings of the 36th International Conference on Machine Learning},
  pages = 	 {3744--3753},
  year = 	 {2019},
  editor = 	 {Chaudhuri, Kamalika and Salakhutdinov, Ruslan},
  volume = 	 {97},
  series = 	 {Proceedings of Machine Learning Research},
  month = 	 {09--15 Jun},
  publisher =    {PMLR},
  url = 	 {https://proceedings.mlr.press/v97/lee19d.html}
}

@inproceedings{transformer,
author = {Vaswani, Ashish and Shazeer, Noam and Parmar, Niki and Uszkoreit, Jakob and Jones, Llion and Gomez, Aidan N. and Kaiser, \L{}ukasz and Polosukhin, Illia},
title = {Attention is all you need},
year = {2017},
isbn = {9781510860964},
publisher = {Curran Associates Inc.},
address = {Red Hook, NY, USA},
booktitle = {Proceedings of the 31st International Conference on Neural Information Processing Systems},
pages = {6000–6010},
numpages = {11},
location = {Long Beach, California, USA},
series = {NIPS'17}
}

@article{lei2016layer,
  title={Layer normalization},
  author={Lei Ba, Jimmy and Kiros, Jamie Ryan and Hinton, Geoffrey E},
  journal={ArXiv e-prints},
  pages={arXiv--1607},
  year={2016}
}

@misc{schulman2017proximalpolicyoptimizationalgorithms,
      title={Proximal Policy Optimization Algorithms}, 
      author={John Schulman and Filip Wolski and Prafulla Dhariwal and Alec Radford and Oleg Klimov},
      year={2017},
      eprint={1707.06347},
      archivePrefix={arXiv},
      primaryClass={cs.LG},
      url={https://arxiv.org/abs/1707.06347}, 
}

@inproceedings{Salamon:UrbanSound:ACMMM:14,
    Address = {Orlando, FL, USA},
    Author = {Salamon, J. and Jacoby, C. and Bello, J. P.},
    Booktitle = {22nd {ACM} International Conference on Multimedia (ACM-MM'14)},
    Month = {Nov.},
    Pages = {1041--1044},
    Title = {A Dataset and Taxonomy for Urban Sound Research},
    Year = {2014}}

@misc{kaggle,
  howpublished = {[EB/OL]},
  note = {\url{https://www.kaggle.com/datasets}},
  title = {Kaggle Dataset Download},
  author = {Jeremy Howard},
  year={2022}
}

@misc{openml,
  howpublished = {[EB/OL]},
  note = {\url{https://www.openml.org}},
  title = {Openml Dataset Download},
  author = {Public},
  year={2022}
}

@misc{uci,
howpublished = {[EB/OL]},
  note = {\url{https://archive.ics.uci.edu/}},
  title = {UCI Dataset Download},
  author = {Public},
  year={2022}
}

@article{ying2024feature,
  title={Feature selection as deep sequential generative learning},
  author={Ying, Wangyang and Wang, Dongjie and Chen, Haifeng and Fu, Yanjie},
  journal={ACM Transactions on Knowledge Discovery from Data},
  volume={18},
  number={9},
  pages={1--21},
  year={2024},
  publisher={ACM New York, NY}
}

@misc{libsvm,
howpublished = {[EB/OL]},
  note = {\url{https://www.csie.ntu.edu.tw/~cjlin/libsvmtools/datasets/}},
  title = {LibSVM Dataset Download},
  author = {Lin Chih-Jen},
  year={2022}
}

@misc{cave,
howpublished = {[EB/OL]},
  note = {\url{https://www.cs.columbia.edu/CAVE}},
  title = {CAVE Dataset Download},
  author = {Nene, Samer A. and Nayar, Shree K. and Murase, Hiroshi},
  year={1996}
}

@misc{osf,
  title={DATASET},
  url={osf.io/qatrz},
  DOI={10.17605/OSF.IO/QATRZ},
  publisher={OSF},
  author={Sajewicz-Radtke, Urszula and Łada-Maśko, Ariadna B and Jurek, Paweł and Olech, Michał and Radtke, Bartosz M},
  year={2024},
  month={Sep}
}

@inproceedings{kbest,
  title={A comparative study on feature selection in text categorization},
  author={Yang, Yiming and Pedersen, Jan O},
  booktitle={Icml},
  volume={97},
  number={412-420},
  pages={35},
  year={1997},
  organization={Nashville, TN, USA}
}

@article{mrmr,
  title={Feature selection based on mutual information criteria of max-dependency, max-relevance, and min-redundancy},
  author={Peng, Hanchuan and Long, Fuhui and Ding, Chris},
  journal={IEEE Transactions on pattern analysis and machine intelligence},
  volume={27},
  number={8},
  pages={1226--1238},
  year={2005},
  publisher={IEEE}
}

@article{lasso,
  title={Regression shrinkage and selection via the lasso},
  author={Tibshirani, Robert},
  journal={Journal of the Royal Statistical Society: Series B (Methodological)},
  volume={58},
  number={1},
  pages={267--288},
  year={1996},
  publisher={Wiley Online Library}
}

@inproceedings{marlfs,
  title={Automating feature subspace exploration via multi-agent reinforcement learning},
  author={Liu, Kunpeng and Fu, Yanjie and Wang, Pengfei and Wu, Le and Bo, Rui and Li, Xiaolin},
  booktitle={Proceedings of the 25th ACM SIGKDD International Conference on Knowledge Discovery \& Data Mining},
  pages={207--215},
  year={2019}
}

@article{mcdm,
  title={Ensemble of feature selection algorithms: a multi-criteria decision-making approach},
  author={Hashemi, Amin and Dowlatshahi, Mohammad Bagher and Nezamabadi-pour, Hossein},
  journal={International Journal of Machine Learning and Cybernetics},
  volume={13},
  number={1},
  pages={49--69},
  year={2022},
  publisher={Springer}
}

@inproceedings{lassonet,
  title={Lassonet: Neural networks with feature sparsity},
  author={Lemhadri, Ismael and Ruan, Feng and Tibshirani, Rob},
  booktitle={International Conference on Artificial Intelligence and Statistics},
  pages={10--18},
  year={2021},
  organization={PMLR}
}

@article{rfe,
  title={Recursive feature elimination with random forest for PTR-MS analysis of agroindustrial products},
  author={Granitto, Pablo M and Furlanello, Cesare and Biasioli, Franco and Gasperi, Flavia},
  journal={Chemometrics and intelligent laboratory systems},
  volume={83},
  number={2},
  pages={83--90},
  year={2006},
  publisher={Elsevier}
}

@article{rra,
  title={Ensemble feature selection: homogeneous and heterogeneous approaches},
  author={Seijo-Pardo, Borja and Porto-D{\'\i}az, Iago and Bol{\'o}n-Canedo, Ver{\'o}nica and Alonso-Betanzos, Amparo},
  journal={Knowledge-Based Systems},
  volume={118},
  pages={124--139},
  year={2017},
  publisher={Elsevier}
}

@INPROCEEDINGS{gains,
  author={Xiao, Meng and Wang, Dongjie and Wu, Min and Wang, Pengfei and Zhou, Yuanchun and Fu, Yanjie},
  booktitle={2023 IEEE International Conference on Data Mining (ICDM)}, 
  title={Beyond Discrete Selection: Continuous Embedding Space Optimization for Generative Feature Selection}, 
  year={2023},
  volume={},
  number={},
  pages={688-697},
  doi={10.1109/ICDM58522.2023.00078}}

@inproceedings{sarlfs,
  title={Efficient Reinforced Feature Selection via Early Stopping Traverse Strategy},
  author={Liu, Kunpeng and Wang, Pengfei and Wang, Dongjie and Du, Wan and Wu, Dapeng Oliver and Fu, Yanjie},
  booktitle={2021 IEEE International Conference on Data Mining (ICDM)},
  pages={399--408},
  year={2021},
  organization={IEEE}
}

@incollection{gfs,
  title={Genetic algorithms in feature selection},
  author={Leardi, Riccardo},
  booktitle={Genetic algorithms in molecular modeling},
  pages={67--86},
  year={1996},
  publisher={Elsevier}
}

@article{mel,
  title={MEL: Efficient Multi-Task Evolutionary Learning for High-Dimensional Feature Selection},
  author={Wang, Xubin and Shangguan, Haojiong and Huang, Fengyi and Wu, Shangrui and Jia, Weijia},
  journal={IEEE Transactions on Knowledge and Data Engineering},
  year={2024},
  publisher={IEEE}
}

@article{forman2003extensive,
  title={An extensive empirical study of feature selection metrics for text classification.},
  author={Forman, George and others},
  journal={J. Mach. Learn. Res.},
  volume={3},
  number={Mar},
  pages={1289--1305},
  year={2003}
}

@article{hall1999feature,
	title={Feature selection for discrete and numeric class machine learning},
	author={Hall, Mark A},
	year={1999},
	publisher={Computer Science, University of Waikato}
}

@inproceedings{yu2003feature,
	title={Feature selection for high-dimensional data: A fast correlation-based filter solution},
	author={Yu, Lei and Liu, Huan},
	booktitle={Proceedings of the 20th international conference on machine learning (ICML-03)},
	pages={856--863},
	year={2003}
}

@incollection{yang1998feature,
	title={Feature subset selection using a genetic algorithm},
	author={Yang, Jihoon and Honavar, Vasant},
	booktitle={Feature extraction, construction and selection},
	pages={117--136},
	year={1998},
	publisher={Springer}
}

@inproceedings{kim2000feature,
  title={Feature selection in unsupervised learning via evolutionary search},
  author={Kim, YeongSeog and Street, W Nick and Menczer, Filippo},
  booktitle={Proceedings of the sixth ACM SIGKDD international conference on Knowledge discovery and data mining},
  pages={365--369},
  year={2000}
}

@article{narendra1977branch,
	title={A branch and bound algorithm for feature subset selection},
	author={Narendra, Patrenahalli M. and Fukunaga, Keinosuke},
	journal={IEEE Transactions on computers},
	number={9},
	pages={917--922},
	year={1977},
	publisher={IEEE}
}

@article{kohavi1997wrappers,
	title={Wrappers for feature subset selection},
	author={Kohavi, Ron and John, George H},
	journal={Artificial intelligence},
	volume={97},
	number={1-2},
	pages={273--324},
	year={1997},
	publisher={Elsevier}
}

@article{sugumaran2007feature,
	title={Feature selection using decision tree and classification through proximal support vector machine for fault diagnostics of roller bearing},
	author={Sugumaran, V and Muralidharan, V and Ramachandran, KI},
	journal={Mechanical systems and signal processing},
	volume={21},
	number={2},
	pages={930--942},
	year={2007},
	publisher={Elsevier}
}

@article{grfg,
  title={Group-wise Reinforcement Feature Generation for Optimal and Explainable Representation Space Reconstruction},
  author={Wang, Dongjie and Fu, Yanjie and Liu, Kunpeng and Li, Xiaolin and Solihin, Yan},
  journal={Proceedings of the 28th ACM SIGKDD international conference on Knowledge discovery and data mining},
  year={2022}
}

@inproceedings{2022traceable,
  title={Traceable Automatic Feature Transformation via Cascading Actor-Critic Agents},
  author={Xiao, Meng and Wang, Dongjie and Wu, Min and Qiao, Ziyue and Wang, Pengfei and Liu, Kunpeng and Zhou, Yuanchun and Fu, Yanjie},
  booktitle={Proceedings of the 2023 SIAM International Conference on Data Mining (SDM)},
  pages={775--783},
  year={2023},
  organization={SIAM}
}

@article{2023traceable,
      title={Traceable Group-Wise Self-Optimizing Feature Transformation Learning: A Dual Optimization Perspective}, 
      author={Meng Xiao and Dongjie Wang and Min Wu and Kunpeng Liu and Hui Xiong and Yuanchun Zhou and Yanjie Fu},
      year={2023},
      eprint={2306.16893},
      archivePrefix={arXiv},
      primaryClass={cs.LG}
}

@misc{ying2023selfoptimizingfeaturegenerationcategorical,
      title={Self-optimizing Feature Generation via Categorical Hashing Representation and Hierarchical Reinforcement Crossing}, 
      author={Wangyang Ying and Dongjie Wang and Kunpeng Liu and Leilei Sun and Yanjie Fu},
      year={2023},
      eprint={2309.04612},
      archivePrefix={arXiv},
      primaryClass={cs.LG},
      url={https://arxiv.org/abs/2309.04612}, 
}

@misc{ying2024unsupervisedgenerativefeaturetransformation,
      title={Unsupervised Generative Feature Transformation via Graph Contrastive Pre-training and Multi-objective Fine-tuning}, 
      author={Wangyang Ying and Dongjie Wang and Xuanming Hu and Yuanchun Zhou and Charu C. Aggarwal and Yanjie Fu},
      year={2024},
      eprint={2405.16879},
      archivePrefix={arXiv},
      primaryClass={cs.LG},
      url={https://arxiv.org/abs/2405.16879}, 
}

@misc{azim2024featureinteractionawareautomated,
      title={Feature Interaction Aware Automated Data Representation Transformation}, 
      author={Ehtesamul Azim and Dongjie Wang and Kunpeng Liu and Wei Zhang and Yanjie Fu},
      year={2024},
      eprint={2309.17011},
      archivePrefix={arXiv},
      primaryClass={cs.LG},
      url={https://arxiv.org/abs/2309.17011}, 
}

@misc{hu2024reinforcementfeaturetransformationpolymer,
      title={Reinforcement Feature Transformation for Polymer Property Performance Prediction}, 
      author={Xuanming Hu and Dongjie Wang and Wangyang Ying and Yanjie Fu},
      year={2024},
      eprint={2409.15616},
      archivePrefix={arXiv},
      primaryClass={cs.LG},
      url={https://arxiv.org/abs/2409.15616}, 
}

@article{kumagai2022few,
  title={Few-shot Learning for Feature Selection with Hilbert-Schmidt Independence Criterion},
  author={Kumagai, Atsutoshi and Iwata, Tomoharu and Ida, Yasutoshi and Fujiwara, Yasuhiro},
  journal={Advances in Neural Information Processing Systems},
  volume={35},
  pages={9577--9590},
  year={2022}
}

@inproceedings{koyama2022effective,
  title={Effective nonlinear feature selection method based on hsic lasso and with variational inference},
  author={Koyama, Kazuki and Kiritoshi, Keisuke and Okawachi, Tomomi and Izumitani, Tomonori},
  booktitle={International Conference on Artificial Intelligence and Statistics},
  pages={10407--10421},
  year={2022},
  organization={PMLR}
}

@inproceedings{liu2021efficient,
  title={Efficient Reinforced Feature Selection via Early Stopping Traverse Strategy},
  author={Liu, Kunpeng and Wang, Pengfei and Wang, Dongjie and Du, Wan and Wu, Dapeng Oliver and Fu, Yanjie},
  booktitle={2021 IEEE International Conference on Data Mining (ICDM)},
  pages={399--408},
  year={2021},
  organization={IEEE}
}

@incollection{biesiada2008feature,
  title={Feature selection for high-dimensional data—a Pearson redundancy based filter},
  author={Biesiada, Jacek and Duch, Wlodzis{\l}aw},
  booktitle={Computer recognition systems 2},
  pages={242--249},
  year={2008},
  publisher={Springer}
}

@article{ding2014identification,
  title={Identification of bacteriophage virion proteins by the ANOVA feature selection and analysis},
  author={Ding, Hui and Feng, Peng-Mian and Chen, Wei and Lin, Hao},
  journal={Molecular BioSystems},
  volume={10},
  number={8},
  pages={2229--2235},
  year={2014},
  publisher={Royal Society of Chemistry}
}

@article{ref_arxiv,
  title   = {Optimizing Aesthetic Perception through Human-AI Teaming for Subtle Dimension Identification in Art Annotation},
  author  = {Wang, Mo and Zhang, Ye and He, Jinlong and Zhou, Yupeng and Li, Niantong and Wang, Jianan and Sun, Yifei and Yin, Minghao},
 journal={IEEE Trans. Learn. Technol.},
  year    = {2026}
}

@misc{fedcaps,
      title={Permutation-Invariant Representation Learning for Robust and Privacy-Preserving Feature Selection}, 
      author={Rui Liu and Tao Zhe and Yanjie Fu and Feng Xia and Ted Senator and Dongjie Wang},
      year={2025},
      eprint={2510.05535},
      archivePrefix={arXiv},
      primaryClass={cs.LG},
      url={https://arxiv.org/abs/2510.05535}, 
}

@article{urbanplanning_rui,
author = {Liu, Rui and Zhe, Tao and Peng, Zhong-Ren and Catbas, Necati and Ye, Xinyue and Wang, Dongjie and Fu, Yanjie},
title = {Urban Planning in the Age of Agentic AI: Emerging Paradigms and Prospects},
year = {2026},
issue_date = {December 2025},
publisher = {Association for Computing Machinery},
address = {New York, NY, USA},
volume = {27},
number = {2},
issn = {1931-0145},
url = {https://doi.org/10.1145/3787470.3787474},
doi = {10.1145/3787470.3787474},
journal = {SIGKDD Explor. Newsl.},
month = dec,
pages = {35–42},
numpages = {8}
}

@misc{catf_rui,
      title={City Editing: Hierarchical Agentic Execution for Dependency-Aware Urban Geospatial Modification}, 
      author={Rui Liu and Steven Jige Quan and Zhong-Ren Peng and Zijun Yao and Han Wang and Zhengzhang Chen and Kunpeng Liu and Yanjie Fu and Dongjie Wang},
      year={2026},
      eprint={2602.19326},
      archivePrefix={arXiv},
      primaryClass={cs.MA},
      url={https://arxiv.org/abs/2602.19326}, 
}
